\def\year{2022}\relax
\pdfoutput=1
\documentclass[letterpaper]{article} % DO NOT CHANGE THIS
\usepackage{aaai22}  % DO NOT CHANGE THIS
\usepackage{times}  % DO NOT CHANGE THIS
\usepackage{helvet}  % DO NOT CHANGE THIS
\usepackage{courier}  % DO NOT CHANGE THIS
\usepackage[hyphens]{url}  % DO NOT CHANGE THIS
\usepackage{graphicx} % DO NOT CHANGE THIS
\usepackage{natbib}  % DO NOT CHANGE THIS AND DO NOT ADD ANY OPTIONS TO IT
\usepackage{caption} % DO NOT CHANGE THIS AND DO NOT ADD ANY OPTIONS TO IT
\DeclareCaptionStyle{ruled}{labelfont=normalfont,labelsep=colon,strut=off} % DO NOT CHANGE THIS
\usepackage{algorithm}
\usepackage{algorithmic}
\usepackage{pdfpages}

\newcommand{\LenVec}[1]{\vert\mathcal{#1}\vert}
\newcommand{\defop}[1]{\mathcal{T}_{\rm{#1}}}
\usepackage{amssymb}
\usepackage{amsmath}
\usepackage{amsthm}
\usepackage{subfigure}
\usepackage{booktabs}
\usepackage{multirow}
\newtheorem{definition}{Definition}
\newtheorem{thm}{Theorem}

\newtheorem{corollary}{Corollary}

\title{Robust Action Gap Increasing with Clipped Advantage Learning}
\author {
    Zhe Zhang,\textsuperscript{\rm 1, \rm 2}
    Yaozhong Gan, \textsuperscript{\rm 1, \rm 2}
    Xiaoyang Tan \textsuperscript{\rm 1, \rm 2}
}
\affiliations {
    \textsuperscript{\rm 1} College of Computer Science and Technology, Nanjing University of Aeronautics and Astronautics\\
    \textsuperscript{\rm 2} MIIT Key Laboratory of Pattern Analysis and Machine Intelligence\\
    \{zhangzhe, yzgancn, x.tan\}@nuaa.edu.cn
}
\begin{document}

\maketitle

\begin{abstract}
Advantage Learning (AL) seeks to increase the action gap between the optimal action and its competitors, so as to improve the robustness to estimation errors. However, the method becomes problematic when the optimal action induced by the approximated value function does not agree with the true optimal action. In this paper, we present a novel method, named clipped Advantage Learning (clipped AL), to address this issue. The method is inspired by our observation that increasing the action gap blindly for all given samples while not taking their necessities into account could accumulate more errors in the performance loss bound, leading to a slow value convergence, and to avoid that, we should adjust the advantage value adaptively. We show that our simple clipped AL operator not only enjoys fast convergence guarantee but also retains proper action gaps, hence achieving a good balance between the large action gap and the fast convergence. The feasibility and effectiveness of the proposed method are verified empirically on several RL benchmarks with promising performance.
\end{abstract}

\section{Introduction}
Many recent studies have shown that (deep) reinforcement learning (RL) algorithms can achieve great progress when making use of regularization, though they may be derived from different motivations, such as robust policy optimization \cite{Schulman15, Schulman17} or efficient exploration \cite{Haarnoja17, Haarnoja18}. According to the reformulation in \cite{vieillard20, Nino20}, Advantage Learning (AL) \cite{bellemare16} can also be viewed as a variant of the Bellman optimality operator imposed by an implicit Kullback-Leibler (KL) regularization between two consecutive policies. And this KL penalty can help to reduce the policy search space for stable and efficient optimization.

Specifically, the AL operator adds a scaling advantage value term to Bellman optimality operator. Besides transformed into an implicit KL-regularized update, this operator can directly increase the gap between the optimal and suboptimal actions, called \textit{action gap}. \cite{bellemare16} shows that increasing this gap is beneficial, and especially a large gap can mitigate the undesirable effects of estimation errors from the approximation function. 

However, a potential problem less studied by previous research is that the advantage term may become a burden if the optimal action induced by the approximated value function does $\textit{not}$ align with the true optimal action. This mismatch is common when there exists under-exploration about the current MDP and would lead to a negative advantage term for the true optimal action at the next iteration. 
Consequently, the AL operator could hinder the value improvement about the true optimal and may lead to suboptimal policies. 
%To investigate this issue, we provide an in-depth analysis on the convergence of the AL operator. 
To investigate this issue, we provide an in-depth analysis on the relationship between advantage term and performance loss bound for the AL operator.
%The theoretical result shows that the advantage term could lead to more convergence errors while increasing the action gap, hence slowing down the value/policy updates.We further illustrate this problem by a classic chain-walk example.
The theoretical result shows that the advantage term could lead to more cumulative errors in performance loss bound while increasing the action gap, hence slowing down the value/policy update. We further illustrate this problem by a classic chain-walk example.

To address the above issue, we present an improved AL algorithm named clipped Advantage Learning (clipped AL). Our key idea can be summarized as "\textit{advantage term should not be added without necessity}" according to the principle of \textit{Occam's razor}.
Intuitively, assume that the optimal action induced by the approximated value function were wrong (which is highly likely at the early stage of the training), the action gap term works just like a regularization imposed on two randomly suboptimal actions and hence it makes no sense to continuously enlarge their gap if it has already been very large. Based on this observation, during AL training we first determine whether the current action gap is too small and only increase this gap if it is below some predefined threshold. This can be easily implemented with a clipping function, and hence we call the resulting method Clipped AL. We show that, with this simple mechanism, we could significantly improve the stability of the AL training by reducing the potential adverse effects when the induced optimal action is wrong.
Besides, clipped AL adopts an adaptive clipping mechanism to adjust the advantage term more reasonably for a robust action gap increasing. From the perspective of implicit regularization, clipped AL can also be viewed as a relaxation on the KL constraints.
We prove that a theoretical balance between fast convergence and large action gap can be achieved by clipped AL. 
%and show that our method can also lead to a smooth difference of action gaps across state space for stable learning. 
Empirical performance on popular RL benchmarks also verifies the feasibility and effectiveness of our clipped AL.

\section{Related Work}

To better understand Advantage Learning, many researchers have tried to analyze and explain the actual effects of action-gap regularity adopted by the AL operator. Farahmand \cite{farahmand11} studied the action gap phenomenon for two-action discounted MDPs and proved that smaller performance loss could be achieved by the problem with a favorable action-gap regularity. Vieillard et al. \cite{vieillard20} drew a connection between an implicit KL regularization with action-gap regularity, which is thought of as beneficial to stable learning. Besides, Seijen et al. \cite{harm19} proposed a hypothesis that a larger difference in the action-gap sizes across the state-space would hurt the performance of approximate RL, which was also supported by strong empirical evidence.

Recent work aims to improve advantage learning mainly from two perspectives. One direction is to extend the idea of AL to the other RL methods. For example, Ferret et al. \cite{Ferret21} connected self-imitation learning (SIL) \cite{sil18} with AL for an optimistic exploration, while by incorporating the AL operator with Retrace \cite{Retrace16}, Kozuno et al. \cite{Grape19} proposed a multi-step version of the AL algorithm. Another direction is to seek a more robust gap-increasing. Conservative valuation iteration (CVI) \cite{kozuno19} achieved a \textit{soft gap-increasing} by replacing $\max$ operators in AL with softmax ones, which could control the trade-off between error-tolerance and convergence rate. Munchausen DQN (MDQN) \cite{vieillard20} also adopted a clipping function on its log-policy term so as to avoid the numerical issue, when implementing the soft gap-increasing.

\section{Preliminaries}

We also formulate the RL problem within the Markov Decision Processes (MDP) framework as commonly considered. Each specific MDP can be modeled as a unique tuple $\mathcal{M}=\langle\mathcal{S},\mathcal{A}, P, r,\gamma\rangle$, where $\mathcal{S}$ and $\mathcal{A}$ denote the state and action space, $P$ is the Markov transition probability function $P:\mathcal{S}\times\mathcal{A}\times\mathcal{S}\rightarrow[0,1]$, $r$ represents the reward function $r:\mathcal{S}\times\mathcal{A}\rightarrow[R_{\min}, R_{\max}]$, and $\gamma$ is the discount factor. The RL agent interacts with the environment following a policy $\pi:\mathcal{S}\times\mathcal{A}\rightarrow[0,1]$\footnote[1]{Note that we may slightly abuse some function notations as the corresponding vector notations in the later, which depends on the context.} 

\subsection{Bellman Operator}
In common to estimate the quality of a policy, the expected discounted cumulative return, denoted by the state value function $V^\pi(s)=\mathbb{E}_{\pi}\left[\sum_{t=0}^\infty\gamma^t r(s_t,a_t)|s_0 = s\right]$, is chosen as the evaluation criterion, where $\mathbb{E}_{\pi}$ represents the expectation over all trajectories $\left(s_0, a_0, r_0, \cdots s_t, a_t, r_t\cdots\right)$ sampled by $\pi$ and $P$. And similarly, the action-state value function is defined as $Q^\pi(s,a)=\mathbb{E}_{\pi}\left[\sum_{t=0}^\infty\gamma^tr(s_t,a_t)|s_0=s,a_0=a\right]$. What an optimal policy aims at is to maximize the value function $V^\pi(s)$ or $Q^\pi(s,a)$ over the space of \textit{non-stationary} and \textit{randomized} policies $\Pi$: $V^*(s)=\sup_{\pi\in\Pi}V^\pi(s)$, and $Q^*(s, a)=\sup_{\pi\in\Pi}Q^\pi(s,a)$.
%\begin{equation}
%	\pi^*(s)=\arg\max_{a\in\mathcal{A}}\sup_{\pi\in\Pi} Q^\pi(s,a)
%\end{equation}
%where the optimal policy satisfies $V^{\pi^*}(s)=\sup_{\pi\in\Pi}V^\pi(s)$. 
And it has been shown that there exists a stationary and deterministic $\pi^*$ that satisfies $V^{\pi^*}=V^*$ and $Q^{\pi^*}=Q^*$ for each $\mathcal{M}$.
%\cite{rltheorybook}.

As we all know, the optimal state-action value function $Q^*$ shared by all the optimal policies satisfies the \textit{Bellman optimality equation}:
\begin{equation}
	Q^*(s, a) = r(s, a) + \gamma\mathbb{E}_{s'\sim P(\cdot|s,a)}\left[\max_{a'}Q^*(s', a')\right]\label{eq:bellman_eq}
\end{equation}
By rewriting Eq.(\ref{eq:bellman_eq}) as the vector form, we can define the \textit{Bellman optimality operator} $\mathcal{T}:\mathbb{R}^{\LenVec{S}\LenVec{A}}\rightarrow\mathbb{R}^{\LenVec{S}\LenVec{A}}$ as:
\begin{equation}
	\mathcal{T}Q\triangleq r+\gamma P V\label{eq:bellman_op}
\end{equation}
where $P \in\mathbb{R}^{\LenVec{S}\LenVec{A}\times\LenVec{S}}, Q\in\mathbb{R}^{\LenVec{S}\LenVec{A}}, V\in\mathbb{R}^{\LenVec{S}}$ and $V(s)=\max_a Q(s,a)$. $\mathcal{T}$ is a contraction operator whose unique fixed point is the optimal action-state function $Q^*$. 

\subsection{Advantage Learning}
In complex tasks, the $Q$ value function is usually approximated by a parameterized neural network $Q_\theta(s,a)$, of which one obvious challenge is its robustness to the estimation errors.
And to mitigate this issue, Advantage Learning (AL) \cite{bellemare16} is proposed to increase the action gap 
, \textit{i.e.}, the difference between the optimal action value and the suboptimal ones, and its operator can be defined as:
\begin{equation}
	\defop{AL}Q(s,a)\triangleq \mathcal{T}Q(s,a)-\alpha\left(V(s)-Q(s,a)\right)\label{eq:AL_op}
\end{equation}

Where the scaling parameter $\alpha\in [0, 1)$. Compared to $\mathcal{T}$, the only modification in the AL operator is the addition of a scaling advantage function $A(s,a)=Q(s,a)-V(s)$ for each state-action pair and $\defop{AL}$ is consistent with $\mathcal{T}$ when $\alpha=0$.
Ideally, $\defop{AL}$ will decrease the value of suboptimal actions (as $A(s,a)<0$), and keep the consistent optimal action value with $\mathcal{T}$ (as $A(s,a^*)=0$). The AL operator has also been proved (Theorem 1 in \cite{bellemare16}) to obtain some critical properties: \textit{optimality-preserving} and \textit{gap-increasing}, which defined as the following:
\begin{definition}[optimality-preserving]
	An operator $\mathcal{T}'$ is optimality-preserving if, for $\forall\;Q_0\in\mathcal{Q}$ and $s\in\mathcal{S}$, letting $Q_{k+1}=\mathcal{T}'Q_k$,
	$$\hat{V}(s)\triangleq\lim_{k\rightarrow\infty}\max_{a\in\mathcal{A}}Q_k(s,a)$$
	exists, is unique, $\hat{V}(s)=V^*(s)$, and for $\forall a\in\mathcal{A}$, 
	$$Q^*(s,a)<V^*(s)\Rightarrow \limsup_{k\rightarrow\infty} Q_k(s,a)<V^*(s)$$
\end{definition}
According to the definition of optimality-preserving, it's suggested that, when using the AL operator, at least one optimal action remains optimal, and all suboptimal actions are still suboptimal.

\begin{figure*}[tb]
	\begin{center}
		\centerline{\includegraphics[width=0.8\linewidth]{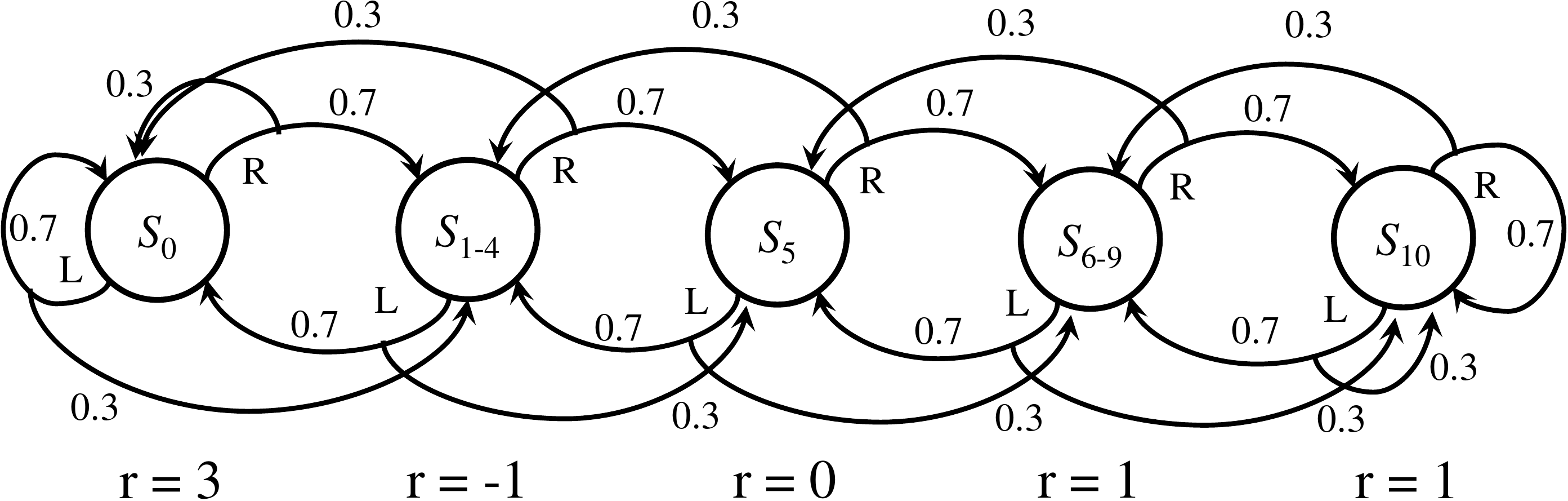}}
		\caption{The 11-state chain-walk example used in \cite{kozuno17}. The optimal policy is to take the left movement regardless of the initial and current states, so that the agent can arrive and stay at the left end state $s_0$ for larger long-horizon rewards.}
		\label{fig:chainwalk}
	\end{center}
\end{figure*}

\begin{definition}[gap-increasing]
	Let $\mathcal{M}$ be a MDP, an operator $\mathcal{T}'$ for $\mathcal{M}$ is gap-increasing if for $\forall Q_0\in\mathcal{Q}, s\in\mathcal{S}, a\in\mathcal{A}$, letting $Q_{k+1}\triangleq\mathcal{T}'Q_k$ and $V_k(s)\triangleq\max_{a'}Q_k(s, a')$,
	$$\liminf_{k\rightarrow\infty}\left[V_k(s)-Q_k(s,a)\right]\geq V^*(s)-Q^*(s,a)$$
\end{definition}
The property of \textit{gap-increasing} implies that the AL operator will enlarge the value difference between the optimal and suboptimal actions than $\mathcal{T}$ does. In fact, Theorem 1 in \cite{kozuno17} shows that the action gaps obtained by $\defop{AL}$ and $\mathcal{T}$ satisfy: $\lim_{k\rightarrow\infty}[V_k(s)-Q_k(s,a)]=\frac{1}{1-\alpha}[V^*(s)-Q^*(s,a)]$.

\section{Performance Loss Bound of AL}
\label{per_analysis}
The additional scaling advantage term in the AL operator contributes to increase the action gap, thereby achieving the robust learning of the value function. 
%However, the advantage value term may also become a burden for the convergence of value function.
However, the advantage value term may also become a burden for the value iteration.
In this section, we will analyze the relationship between the advantage term and the performance loss bound of the AL operator, which leads to our motivation on improving the AL operator.

Starting with an arbitrary initial action-state value function $Q_0\in \mathcal{Q}$, we can obtain an action-state value function sequence  $\left\{Q_k\right\}_{k=0}^K$ by iteratively applying the AL operator $\defop{AL}$, \textit{i.e.}, $Q_{k+1}=\defop{AL}Q_k$. And we get the corresponding state value function sequence $\left\{V_k\right\}_{k=0}^K$ by following the definition: $V_k(s)\triangleq\max_a Q_k(s,a)$. Because $\defop{AL}$ is \textit{optimality-preserving}, we know that the state value function sequence will converge to the optimal one, \textit{i.e.}, $\lim_{k\rightarrow\infty}V_k=V^*$. The greedy policy induced by the $k$-th state value function $V_k$ is defined as: $\pi_{k+1}(s)=\arg\max_a\left[r(s,a)+\gamma\mathbb{E}_{s'|s,a}\left[V_k(s')\right]\right]$, 
and then the $\ell_\infty$-norm performance loss bound of state value function  of the induced policy satisfies the following result\footnote[2]{we analyze the convergence error, because the sequence $\left\{V_k\right\}^K_{k=0}$ must converge to the optimal, while $\left\{Q_k\right\}^K_{k=0}$ may not, according the definition of optimality-preserving.}: (proof in Appendix A.1)

\begin{thm}
	Assume the optimal policy $\pi^*$ and its state value function $V^*$, and $\forall \pi\in\Pi, \Vert V^\pi\Vert_\infty\leq V_{\max}$, let $\Delta^{\pi^*}_{k}\in\mathbb{R}^{\LenVec{S}}$ and each entry is defined as :$\Delta_k^{\pi^*}(s) = V_{k}(s)-Q_{k}(s,\pi^*(s))$, then we have:
	\vspace{3pt}
	\begin{align}
		&\Vert V^*-V^{\pi_{K+1}}\Vert_\infty\nonumber\\
		&\leq\frac{2\gamma}{1-\gamma}\left[2\gamma^KV_{\max}+\alpha\sum_{k=0}^{k=K-1}\gamma^{K-k-1}\Vert\Delta_k^{\pi^*}\Vert_\infty\right]\nonumber
	\end{align}\label{thm_error}
\end{thm}

%Comparing the error bound in Theorem \ref{thm_error} with the similar one of Bellman optimality operator \cite{Farahmand10}, we can see that $\defop{AL}$ would introduce an extra discounted sum of the norms of errors, which slows down the convergence to the optimal policy. 
Comparing the result in Theorem \ref{thm_error} with the similar one of Bellman optimality operator \cite{Farahmand10}, we can see that $\defop{AL}$ would accumulate an extra discounted error into the performance loss bound in the case that a non-zero $\Delta_k^{\pi^*}$ occurs at any update step. And the additional cumulative errors will further lead to a slower convergence to the optimal value function.

Recall the definition $\Delta_k^{\pi^*}(s) = V_{k}(s)-Q_{k}\left(s,\pi^*(s)\right)$, we know it's a non-negative vector ($\Delta^{\pi^*}_{k}\geq 0$) and $-\Delta^{\pi^*}_{k}(s)$ represents the estimated advantage value of the true optimal action at state $s$. When the optimal action induced by the iterative value function does not agree with the true optimal action, \textit{i.e.}, $\pi^*(s)\neq\arg\max_{a\in\mathcal{A}}Q_k(s,a)$ at some timesteps, a positive discounted error $\gamma^{K-k-1}\Vert\Delta^{\pi^*}_k\Vert_\infty>0$ will be accumulated in the performance loss bound. 
%In other words, unless the induced greedy policy keeps consistent with the optimal policy over all the iterations, \textit{i.e.}, $\pi_1=\cdots=\pi_{K+1}=\pi^*$, $\defop{AL}$ would cause larger convergence errors. 
In other words, unless the induced greedy policy keeps consistent with the optimal policy over all the iterations, \textit{i.e.}, $\pi_1=\cdots=\pi_{K+1}=\pi^*$, $\defop{AL}$ would cause larger performance loss bound than $\mathcal{T}$ does. 
However, it's impossible to guarantee this ideal condition in practice, especially because of the under-exploration in complex tasks and the estimation error that existed in the function approximator. So the AL operator also suffers from slower value convergence (\textit{i.e.}, larger performance loss ) while obtaining larger action gaps.

In summary, we show that increasing the action gap by the advantage term is not always a beneficial choice, especially when the induced optimal action is not consistent with the true optimal one. Because the advantage term in this case may also introduce more errors into the state value function, leading to the slow convergence.

\begin{figure*}[tb]
	\centering
	\subfigcapskip=0pt
	\subfigure[]{
		\centering
			\includegraphics[width=0.40\textwidth]{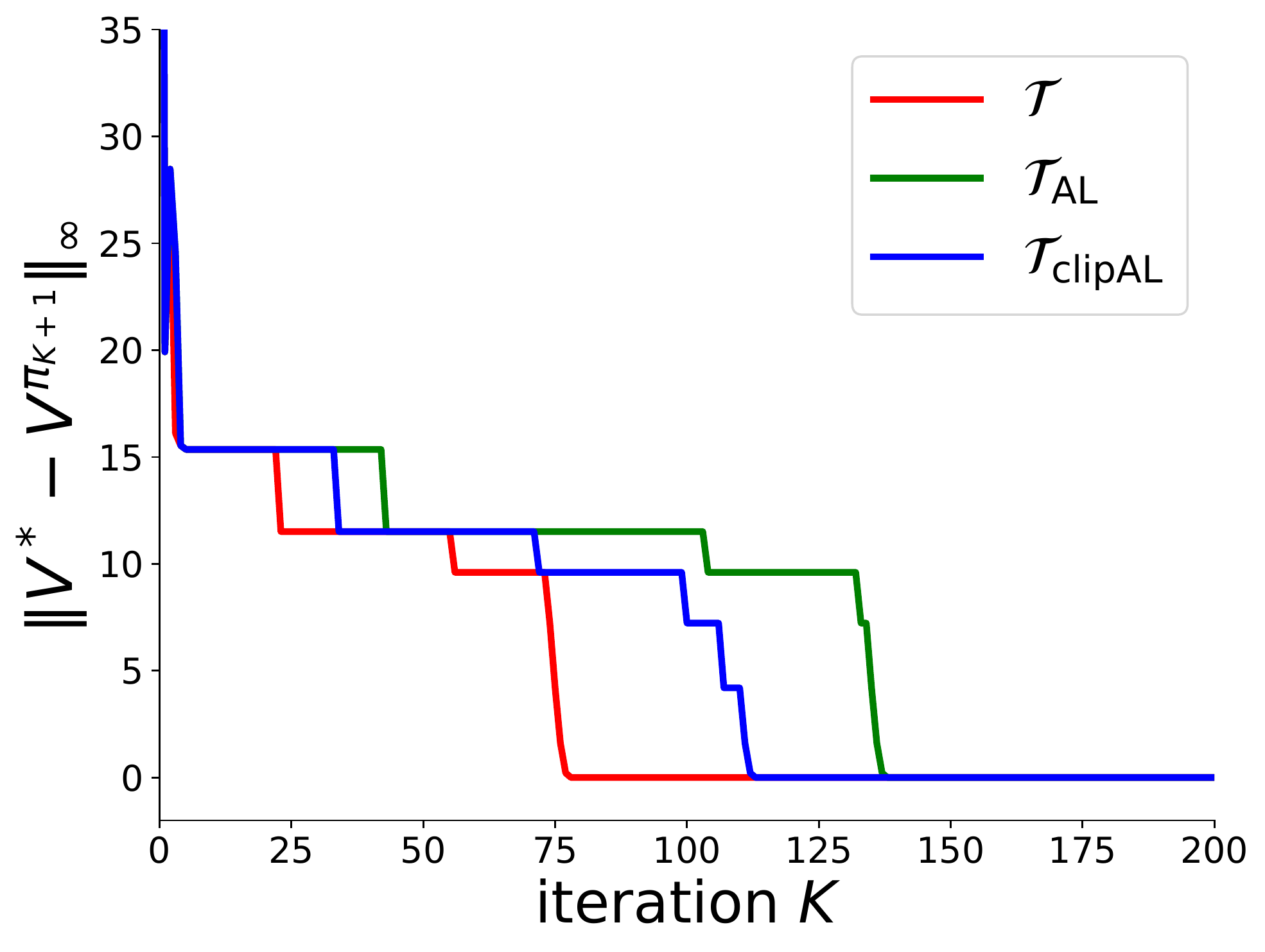}
			\label{fig:cw_V_con}
	}
	\subfigure[]{
		\centering
			\includegraphics[width=0.27\textwidth]{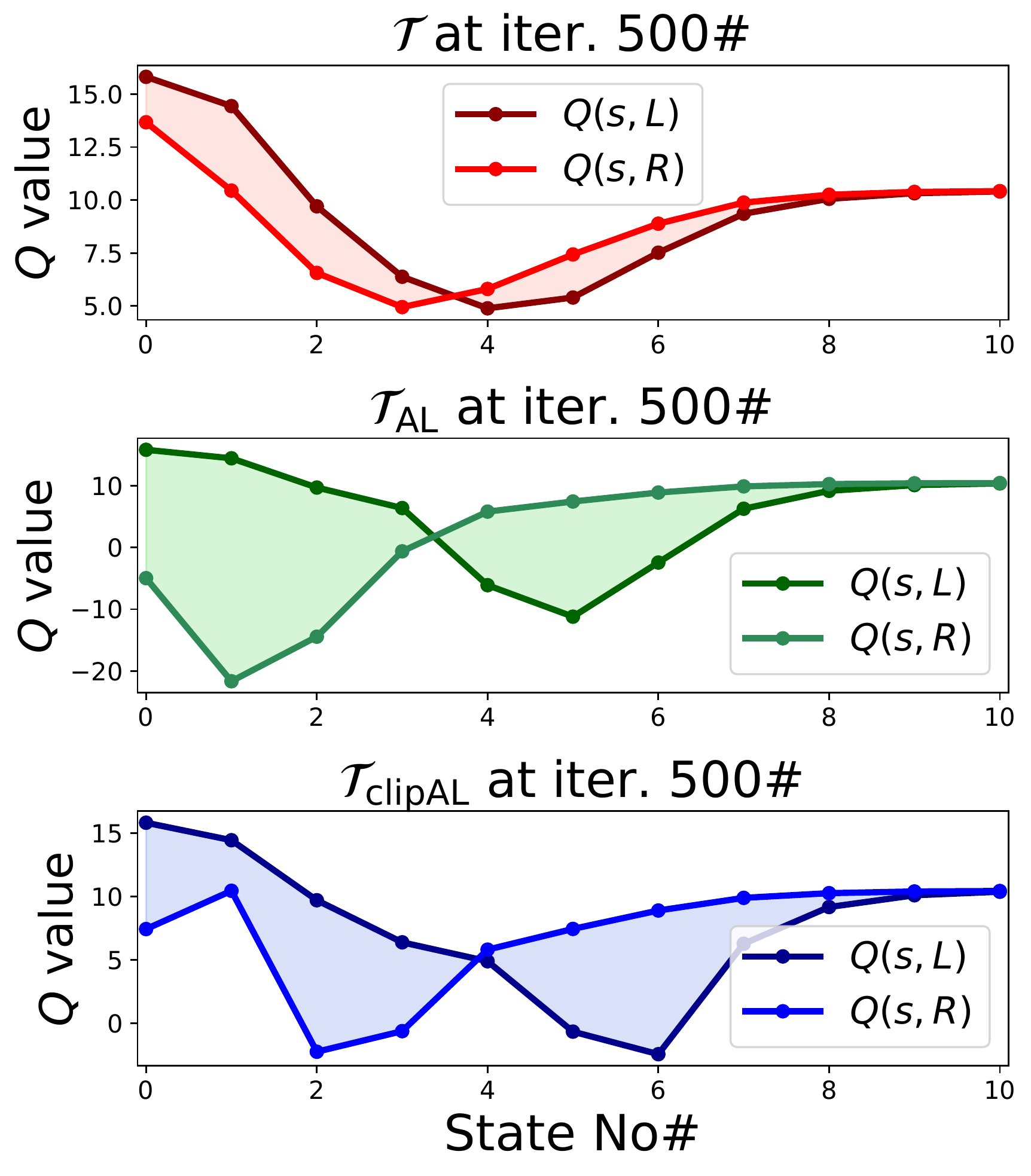}
			\label{fig:early_Q_iter}
	}
	\subfigure[]{
		\centering
			\includegraphics[width=0.27\textwidth]{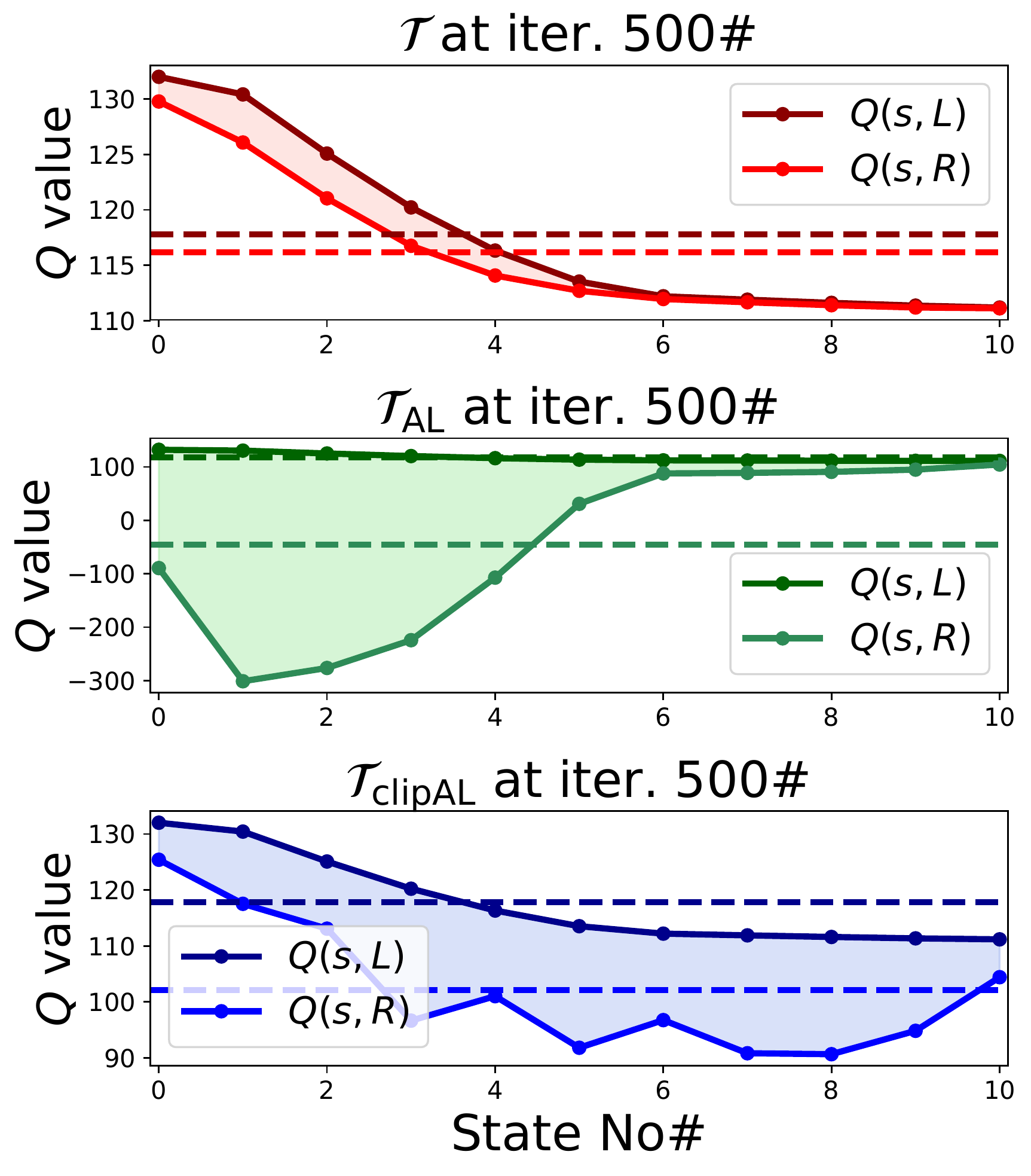} 
			\label{fig:cw_Q_iter}
	}
	\caption{
	Numerical experiments on 11-state chain-walk. \textbf{(a)}: Performance loss bound. The induced policies by $\mathcal{T}$, $\defop{AL}$ and $\defop{clipAL}$ reach the optimal one after 78, 138 and 113 iterations respectively; \textbf{(b-c)}: $Q$ value at 10-th and 500-th iteration. The solid lines depict the $Q$ value of both actions at each state. The dashed lines show the averaged $Q$ value of both actions over all states and the distance between them represents the mean action gap. \textbf{(a)} and \textbf{(c)} indicate that our $\defop{clipAL}$ can speed up the policy convergence than $\defop{AL}$ (though slower than $\mathcal{T}$), and still maintain a larger mean action gap ($15.71$) than $\mathcal{T}$ ($1.65$), so as to achieve the balance between convergence speed and action gap.
	}
	\label{fig:cw_result} 
\end{figure*}

\textbf{11-State Chain-Walk}. We further illustrate this adverse effect with the chain-walk example shown in Figure \ref{fig:chainwalk}. The agent can move either left or right at each state and would be transitioned to the state in the intended direction with probability 0.7, while to the state in the opposite direction with probability 0.3. At both ends of the chain, attempted movement to outside of the chain results in staying at the ends. The agent gets $0$ reward once reaching the middle state ($s_5$). If the agent moves to the right side of the chain ($s_6$-$s_{10}$), it can get $1$ reward, otherwise get $-1$ reward on the left side of this chain ($s_1$-$s_4$) except $3$ reward on the left end ($s_0$). 

Assume every episode will start from state $s_5$, according to the definition, we know that the optimal policy is to implement the \textit{'left'} action at all states. We denote the $Q$ value of \textit{'left'}(\textit{'right'}) action as $Q(s, L)$ ($Q(s, R)$) and initiate a $Q$-table in which $Q(s, R)=Q(s, L)=0$ for all states. Then with a perfect environment model, the $Q$-table will be updated using $\mathcal{T}$ and $\defop{AL}$ respectively. 

%Figure \ref{fig:cw_V_con} shows the convergence of state value function of the induced policy. 
Figure \ref{fig:cw_V_con} shows the performance loss of state value function of the induced policy. 
We can see that it spends more time for $\defop{AL}$ (iteration 138) to achieve the optimal state value function than $\mathcal{T}$ does (iteration 78). This is because a suboptimal policy would be learned at the early update iterations. As shown in Figure \ref{fig:early_Q_iter}, at the beginning of iterations (iteration 10), the induced greedy policy will take the suboptimal (\textit{'right'}) action at state $s_4$-$s_{10}$ due to the larger immediate reward at the right side of the chain. And according to our analysis in Theorem \ref{thm_error}, this suboptimal policy would accumulate more errors in $\Vert V^*-V^{\pi_{K+1}}\Vert_\infty$, leading to a slower convergence. Even though a larger mean action gap can be achieved by $\defop{AL}$ ($163.33$) than $\mathcal{T}$ ($1.65$) after converging to the optimal policy as illustrated in Figure \ref{fig:cw_Q_iter}. More experimental details and results about this chain-walk example can be found in Appendix B.1. 

\section{Clipped Advantage Learning}
Based on the observation in Sec.\ref{per_analysis}, we present a novel AL-based method (named \textit{clipped AL}) in this section, which adds the advantage term more reasonably through a clipping mechanism and also prove its some critical properties. 
 
\subsection{Methods}
Besides the robustness benefited from the larger action gap, the AL operator can also cause a slower convergence due to the blind action gap increasing by the advantage term.
%a slower convergence may also be caused by the AL operator because of the blind action gap-increasing.
To mitigate this issue, one intuitive method is to add the advantage term conditionally based on the necessity of increasing the action gap, rather than doing this for all state-action pairs like AL does. 
%To mitigate this issue, one intuitive method is to judge whether it's necessary to increase the action gap of a certain state-action pair and avoid further 
So we propose the Clipped Advantage Learning (clipped AL) operator as following:
\begin{align}
	&\defop{clipAL}Q(s,a)\nonumber\\
	&\triangleq\mathcal{T}Q(s,a)-\alpha(V(s)-Q(s,a))\cdot\mathbb{I}\left[\frac{Q(s,a)-Q_{l}}{V(s)-Q_{l}}\geq c\right]\label{eq:clippedAL1}
\end{align}

where $\mathbb{I}\left[\cdot\right]$ is the indicator function that equals to 1 if the condition is satisfied, otherwise returns 0. And $c\in\left(0,1\right)$ denotes the clipping ratio coefficient. $Q_l$ is a lower bound of $Q$ value such that $\frac{Q(s,a)-Q_{l}}{V(s)-Q_{l}}\geq 0$. 
This operator can also be rewritten as a more direct form:
\begin{equation}	
	\defop{clipAL} =
	\begin{cases} 
		\defop{AL}, & \mbox{if } Q(s,a)-Q_l\geq c(V(s)-Q_l)\\
		\mathcal{T}, & \mbox{otherwise}
	\end{cases}\label{eq:clippedAL2}
\end{equation}

The motivation behind the clipped AL can be summarized as \textit{"advantage term should not be added without necessity"}. According to the definition in Eq.(\ref{eq:clippedAL2}), the clipped AL is designed to increase the action gap by implementing $\defop{AL}$ if and only if the $Q$ value of suboptimal actions exceeds a certain threshold and gets close to the corresponding $V$ value, or otherwise, it will keep consistent with the Bellman optimality operator $\mathcal{T}$. 
%The motivation behind the clipped AL in Eq.(\ref{eq:clippedAL2}) is to increase the action gap by implementing $\defop{AL}$ if and only if the $Q$ value of suboptimal actions exceeds a certain threshold. Otherwise, the clipped AL operator will keep consistent with the Bellman optimality operator $\mathcal{T}$. Specifically, this form can be summarized as \textit{“advantage term should not be added without necessity”}. 
On the one hand, $\defop{clipAL}$ can still maintain an enough action gaps by the additional advantage term when suboptimal state-action values approach to the optimal one. On the other hand, if an appropriate gap has already existed, it can achieve a larger value improvement without the advantage term in the next iteration. And eventually, $\defop{clipAL}$ is expected to reach a balance between large action gaps and fast convergence.

\begin{corollary}
	The clipped AL operator $\defop{clipAL}$ satisfies the both conditions in Theorem 1 in \cite{bellemare16} and then is both optimality-preserving and gap-increasing, 
\end{corollary}

The above Corollary implies that $\defop{clipAL}$ can still keep both optimality-preserving and gap-increasing like $\defop{AL}$. and will eventually yield an optimal greedy policy when the Q value can be represented exactly. This clipping mechanism is beneficial in the case that the estimated value of the true optimal action $Q_k(s,\pi^*(s))$ is much less than the estimated optimal value $V_k(s)$, because it would omit the negative advantage value $Q_k(s,\pi^*(s))-V_k(s)$ so as to achieve a larger improvement on $Q_{k+1}(s,\pi^*(s))$ in the next iteration, and then help the induced optimal action to align with the true optimal action faster.
Note that instead of a fixed $Q$ value threshold, we choose a fixed $Q$ value ratio $c$ as the clipping threshold to adjust the advantage value term adaptively according to the varying scale of action value.

\subsection{Balance between Large Action Gap and Fast Convergence}

Recall the 11-state chain-walk example in Figure \ref{fig:chainwalk}. We know that, despite increasing the action gap, $\defop{AL}$ may also lead to a slow convergence to the optimal value function because of the mismatch between induced and true optimal action. We also implement $\defop{clipAL}$ on chain-walk example with the same settings and show the results in Figure \ref{fig:cw_result}. We can see that, comparing with $\defop{AL}$, our $\defop{clipAL}$ can obtain a faster achievement to the optimal policy (iteration 113), \textit{i.e.}, $V^{\pi_{K+1}}=V^*$ (shown in Figure \ref{fig:cw_V_con}). This result is intuitive because $\defop{clipAL}$ would clip the unnecessary advantage term $-\Delta_k^{\pi^*}$, reducing the cumulative errors in performance loss bound. Meanwhile, $\defop{clipAL}$ can also maintain a larger action gap ($15.71$) than $\mathcal{T}$ ($1.65$) as shown in Figure \ref{fig:cw_Q_iter}, which keeps its robustness. 

Specifically, for $\forall (s,a)\in\mathcal{S}\times\mathcal{A}$, we define its action gap as following:
\begin{equation}
	G(s,a) = \liminf_{k\rightarrow\infty}\left[V_k(s) - Q_k(s,a)\right]
\end{equation}

where $\left\{Q_k\right\}_{k=0}^\infty$ and $\left\{V_k\right\}_{k=0}^\infty$ are the corresponding value functions \textit{w.r.t} any operator and $V_k(s)=\max_{a\in\mathcal{A}}Q_k(s,a)$. And the action gap obtained by the above three operators satisfies the conclusion in Theorem\ref{them_ag}. 

\begin{thm}
	For $\forall s\in\mathcal{S}, a\in\mathcal{A}$, we define its action gap from $\mathcal{T}$, $\defop{AL}$, and $\defop{clipAL}$ by $G^*(s,a)$, $G_{\rm{AL}}(s,a)$, and $G_{\rm{clipAL}}(s,a)$. Let $Q^*$ and $V^*$ represent the optimal state (action) value functions, then $G^*(s,a)=V^*(s)-Q^*(s,a)$ and these action gaps satisfy:
	$$G^*(s,a)\leq G_{\rm{clipAL}}(s,a)\leq G_{\rm{AL}}(s,a)$$
	And when $Q_l \leq \min_{s,a}\frac{Q^*-\alpha V^*}{1-\alpha}$, $G_{\rm{clipAL}}(s,a)=G_{\rm{AL}}(s,a)$ if the clipping ratio satisfies:
	$$c\leq\min_{s,a}\frac{Q^*-\alpha V^*-(1-\alpha)Q_l}{(1-\alpha)(V^*-Q_l)}$$\label{them_ag}
\end{thm}

This theorem implies that the action gap of $\defop{clipAL}$ is somewhere between the action gaps of both $\mathcal{T}$ and $\defop{AL}$ and finally depends on the clipping ratio $c$. So these results and conclusions support the goal of our clipped AL: achieve a balance between the large action gaps and fast convergence.

\section{Experiment}\label{title:experiment}
To further verify the feasibility and effectiveness of the proposed clipping mechanism applied in the family of Advantage Learning algorithms, we evaluate and compare the performance of our method on several popular RL benchmarks, such as the MinAtar \cite{MinAtar19}, PLE \cite{PLE16} and Atari \cite{ALE13} .

\subsection{Experimental Setup}\label{title:implement}

\subsubsection*{Implementation.} We conduct the MinAtar and PLE experiments mainly based on the \textit{Explorer} framework \cite{Explorer}, and the Atari experiments based on \textit{APE-X} framework \cite{APE-X18}. 
And due to the paper space limit, the results on Atari tasks will be provided in Appendix B.3. All the implementations are run on a computer with an Intel Xeon(R) CPU, 64GB of memory and a GeForce RTX 2080 Ti GPU.
	
When implementing our \textit{clipped AL}, instead of an enough lower bound $Q_l$, we choose a proper value: $Q_l=\frac{1-\gamma^H}{1-\gamma}R_{\min}$, where $R_{\min}$ is the minimum reward for each step. This choice is the least discounted sum of rewards for a $H$-length trajectory.
Although $\frac{Q(s,a)-Q_{l}}{V(s)-Q_{l}} < 0$ may still happen during the training process due to approximation error at certain timesteps, it equals to implement $\mathcal{T}$ in this case. And we know that the fixed point of $\mathcal{T}$ must be greater than or equal to $\frac{1-\gamma^H}{1-\gamma}R_{\min}$, so the ratio will still be non-negative after some iterations. Except the clipping ratio $c$, we select the same hyperparameter used in AL method and more details about the settings can be found in Appendix B.2.

\subsubsection*{Baselines.} To verify our method sufficiently, we compare the clipped AL with several popular baselines as following:
\begin{itemize}
	\item \textbf{AL}: the original Advantage Learning algorithm \cite{bellemare16}, which is the basic method we modify. And we adopt the recommended $\alpha=0.9$;
	\item \textbf{DQN}: the vanilla DQN \cite{Mnih15}, a famous baseline commonly used in discrete-action environment;
	\item \textbf{MDQN}: the Munchausen DQN \cite{vieillard20}, which is a state-of-the-art non-distRL algorithm. And we follow its hyperparameter suggestions: $\alpha=0.9$ (Munchausen scaling term), $\tau=0.03$ (entropy temperature), and $l_0=-1$ (clipping value);
	\item \textbf{Soft-DQN($\tau$)}: the vanilla DQN with maximum entropy regularization, \textit{i.e.}, the discrete-action version of Soft Actor-Critic (SAC) \cite{haarnoja18b}, we set the same temperature parameter $\tau=0.03$ with MDQN;
\end{itemize}

\begin{figure*}[tb]
	\centering
	\subfigbottomskip=1pt
	\subfigure{
			\includegraphics[width=0.3\linewidth]{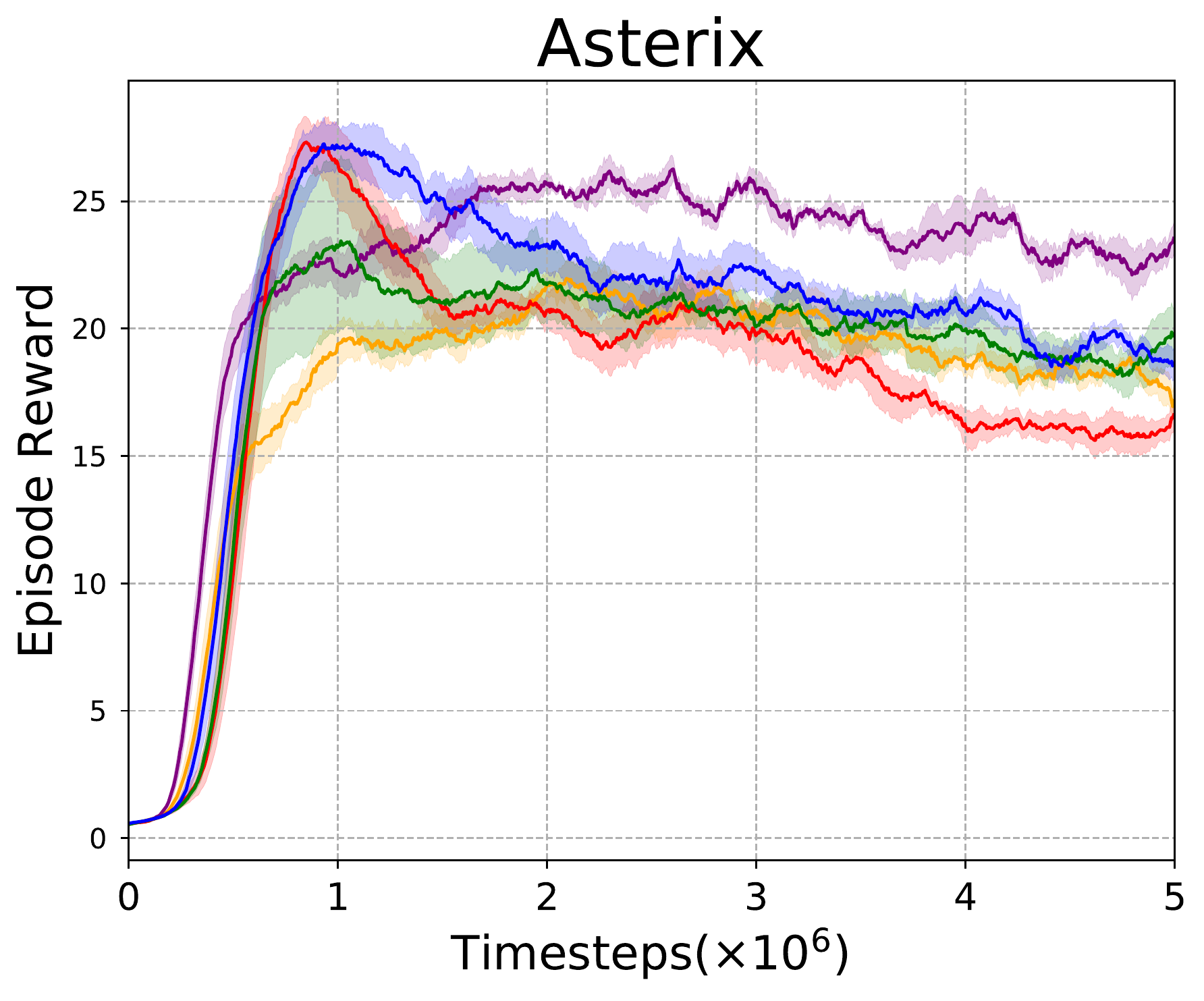}
    }
	\subfigure{
			\includegraphics[width=0.3\linewidth]{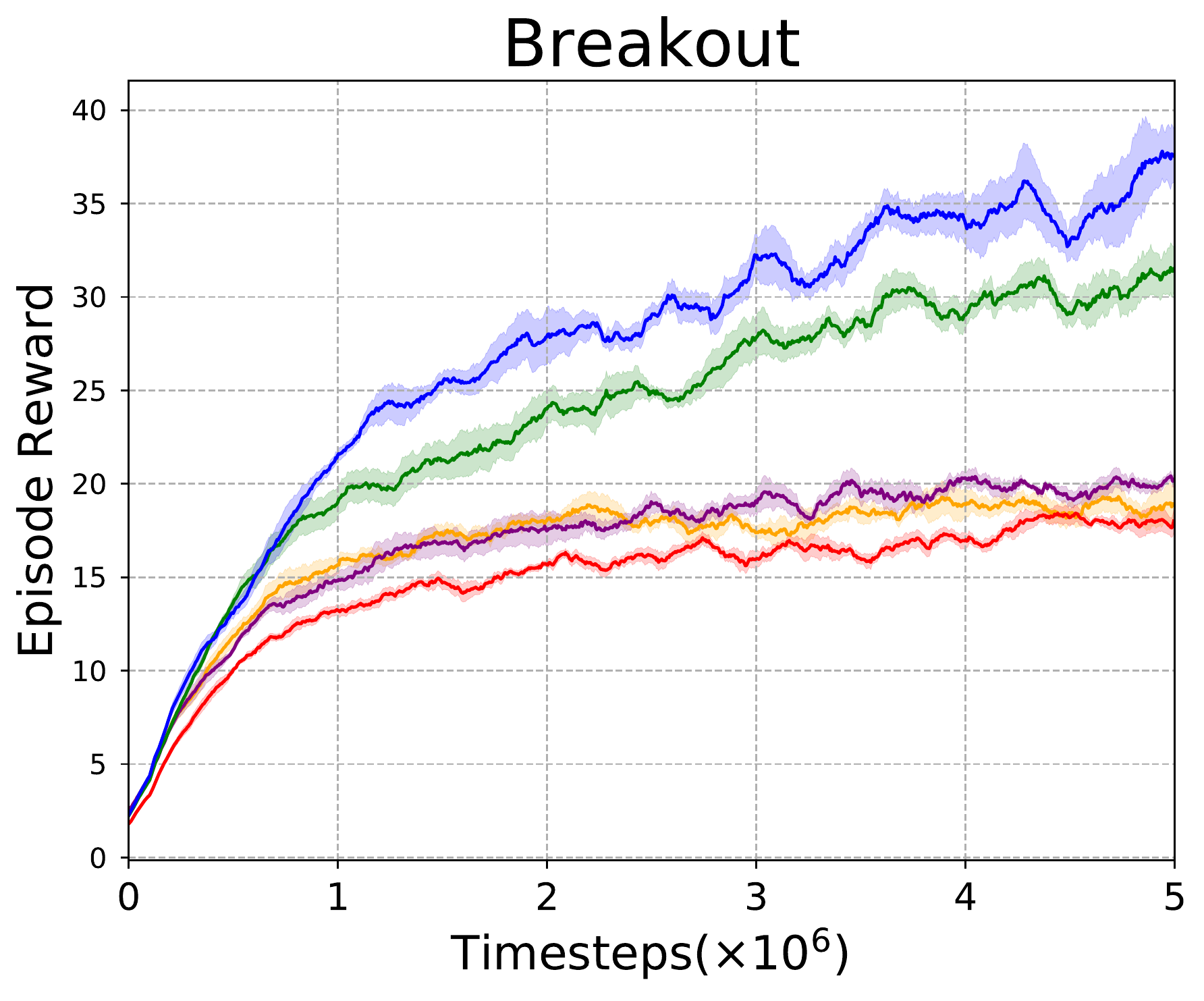}
	}
	\subfigure{
			\includegraphics[width=0.3\linewidth]{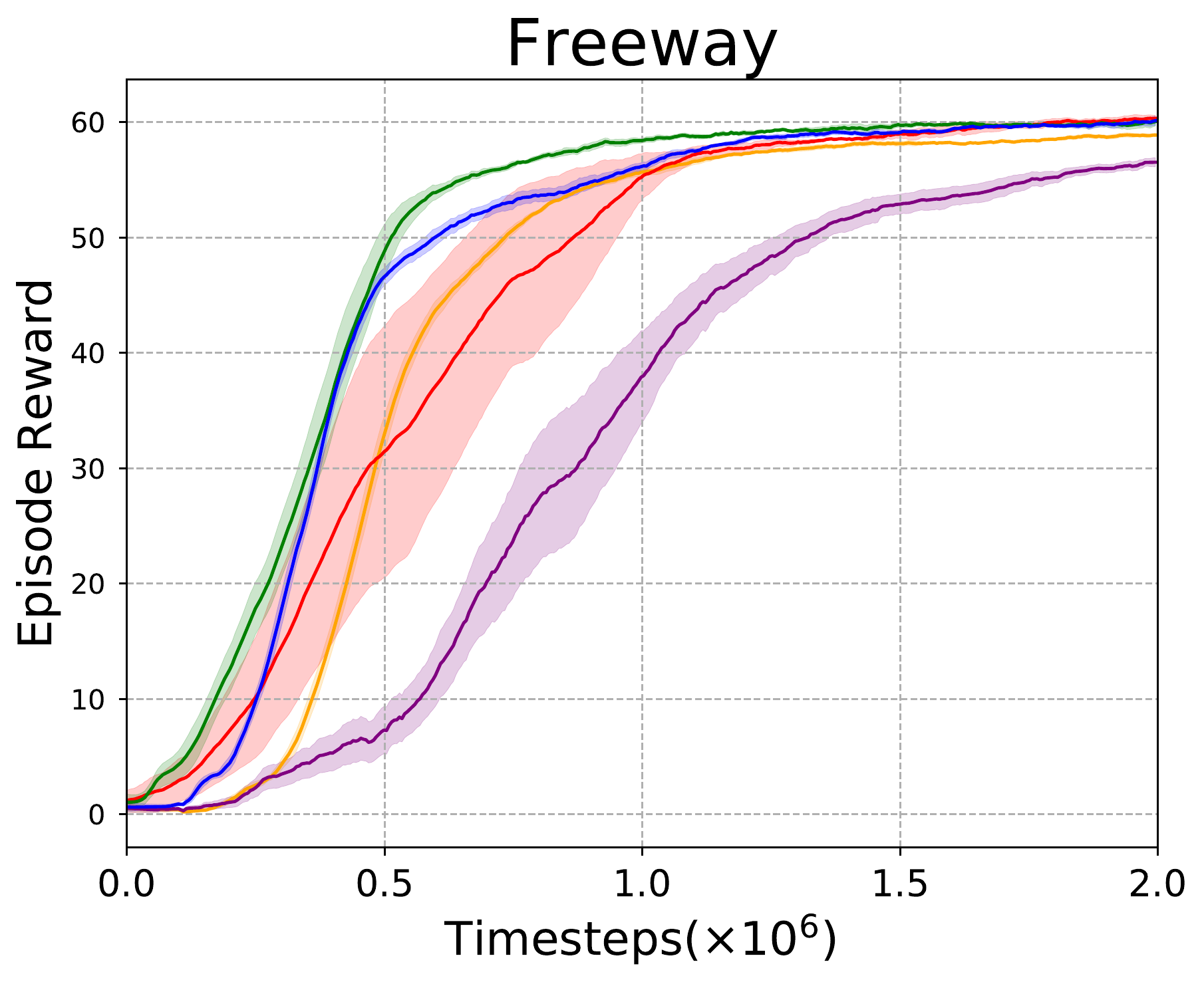}
	}
	\\
	\subfigure{
			\includegraphics[width=0.30\linewidth]{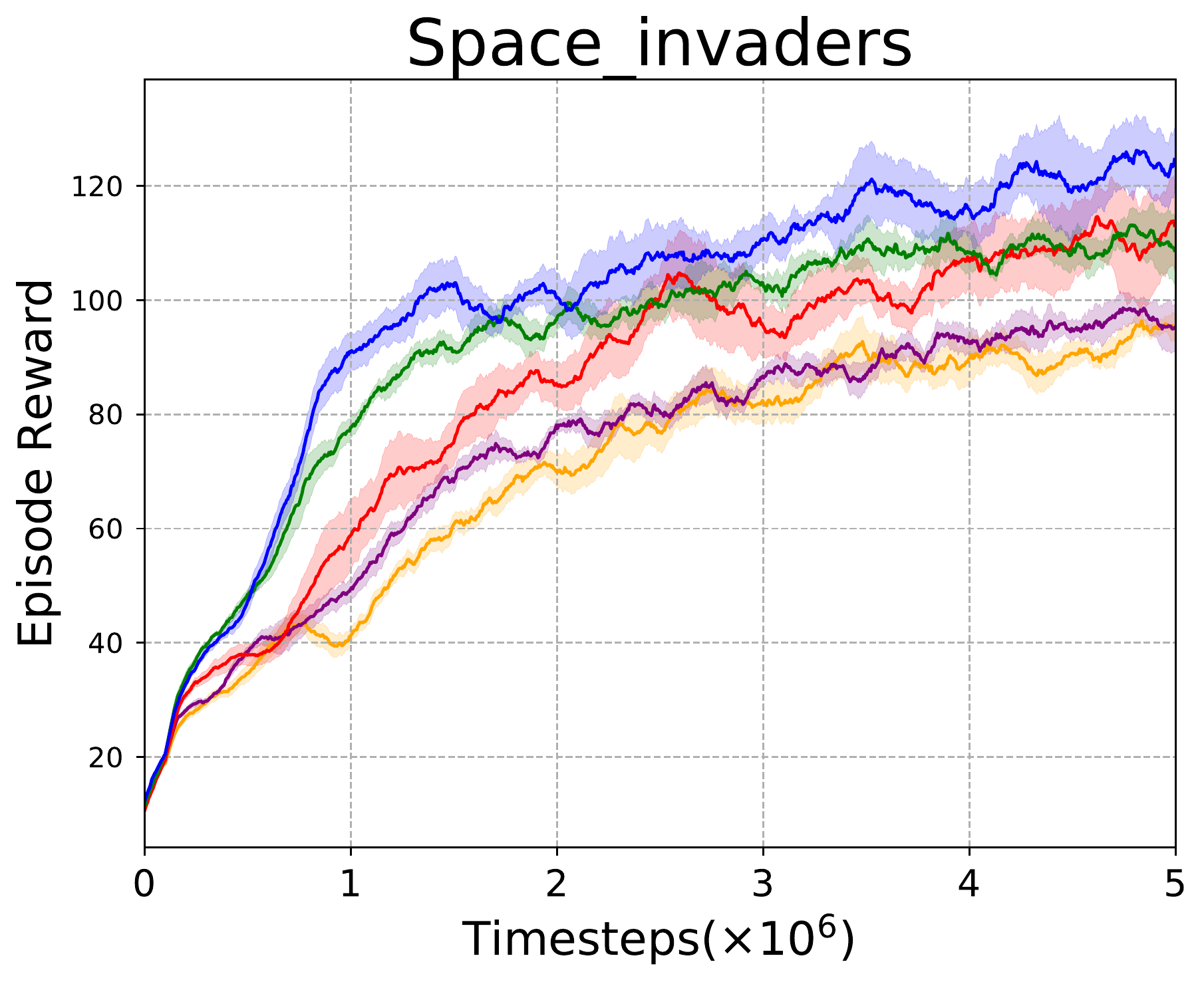}
	}
	\subfigure{
			\includegraphics[width=0.30\linewidth]{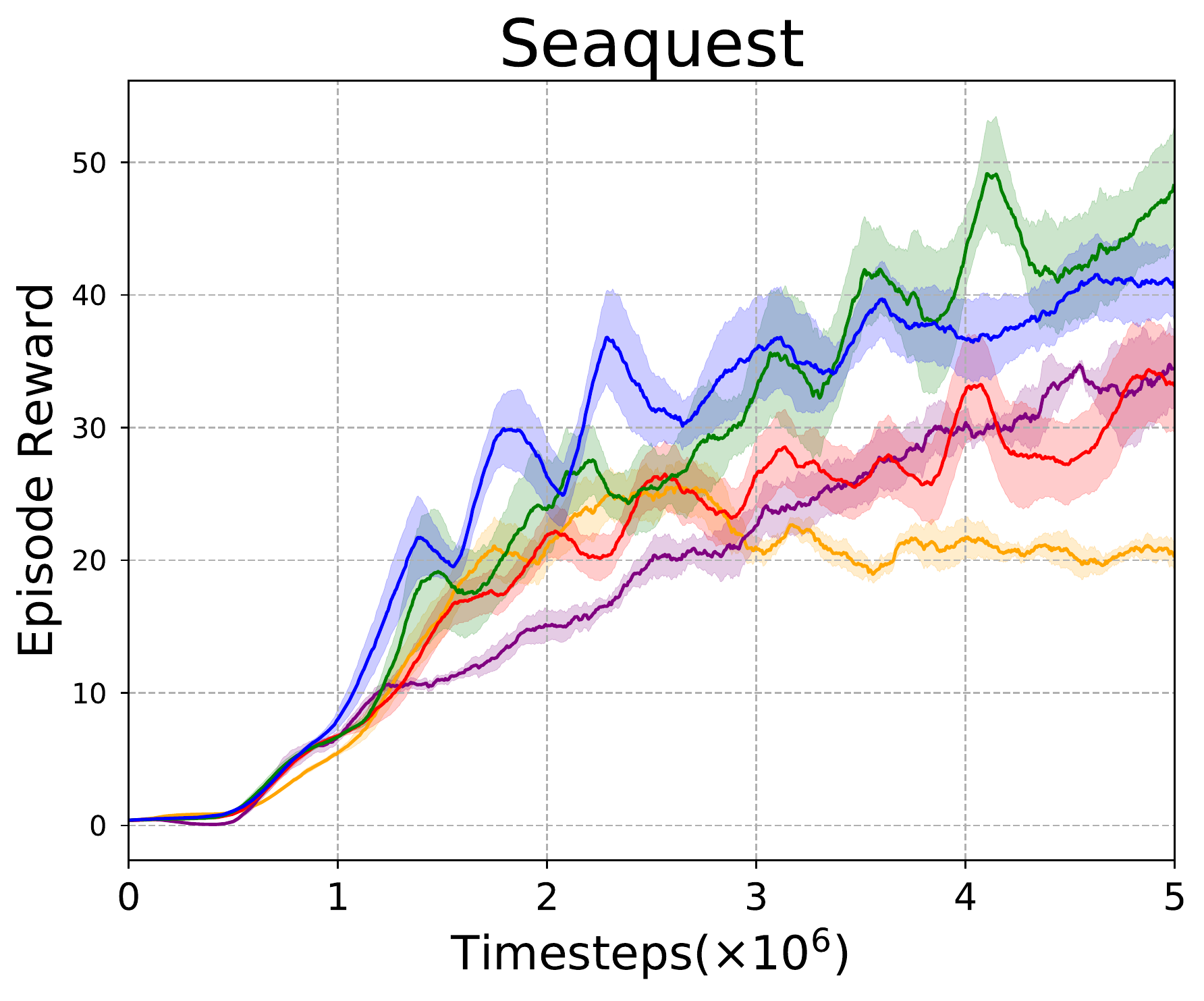}
	}
	\subfigure{
			\includegraphics[width=0.30\linewidth]{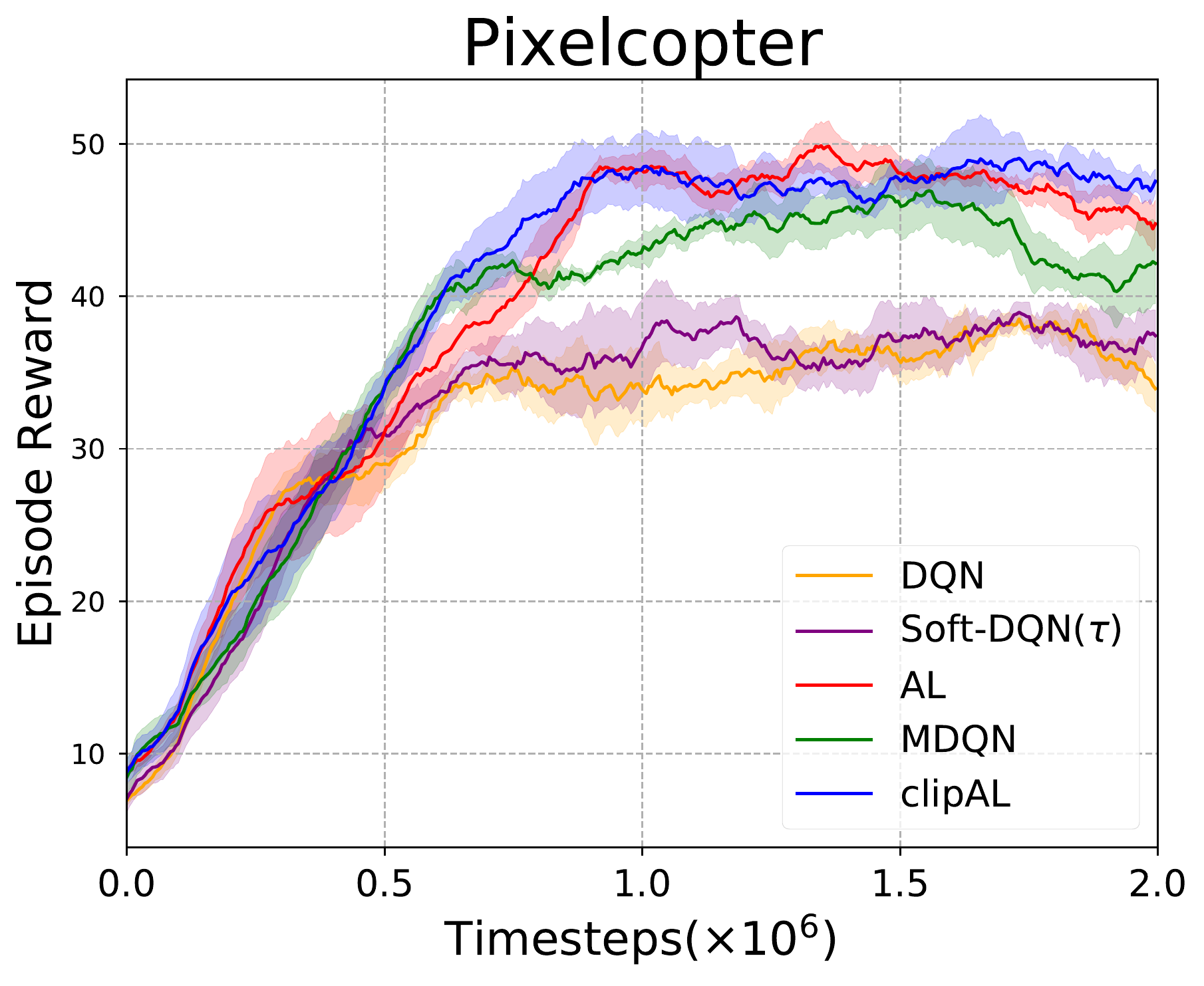}
	}
	\caption{Evaluation episode reward comparison between our clipped AL algorithm and the chosen baselines on six benchmark tasks. All the results are averaged over five random seeds, with the shade area corresponding to one standard error.}
	\label{fig:eff}
\end{figure*}

\subsubsection*{Evaluation.} As for the evaluation scenarios, we select five MinAtar tasks (\textit{Asterix, Breakout, Freeway, Space-invaders}, and \textit{Seaquest}) and one PLE task (\textit{Pixelcopter}). We separate the evaluation from the training process, and conduct policy evaluation per 5000 timesteps. Specifically, we measure the policy performance by the mean reward of 10 evaluation episodes, and all the performance results are averaged over 5 random seeds. 

To further quantify the performance improvement, we adopt the "baseline-normalized" score as the metric. At each evaluation, the score is the undiscounted sum of rewards, averaged over the last 5 evaluations. The normalized score is then $\frac{a-r}{\vert b-r\vert}$, with $a$ the score of the compared algorithm, $b$ the score of the baseline, and $r$ the score of a random policy.

\subsection{Effectiveness of Clipping Mechanism}

\subsubsection*{Performance Improvement.} We firstly validate the effectiveness of our clipped AL. Figure \ref{fig:eff} shows the performance comparison between our method and the baselines mentioned above. We can see that our clipped AL performs better significantly than the AL operator over 5 tasks except Freeway task. Though both methods have a similar final performance on Freeway task, our clipped AL still has a higher sample efficiency before the final convergence. Even comparing with MDQN and Soft-DQN($\tau$), our method is also competitive and achieve the best performance on Breakout, Space invaders and Pixelcopter tasks. 
We compute the 'DQN-normalized' score for the other 4 methods and Table \ref{table:dqn-normalized} depicts the quantitative results.
We can see from it that our clipped AL achieves around 45.73 \% averaged performance improvement, which is better than the rest baselines
and more than double times than the original AL method especially. All the results can verify our clipping mechanism does improve the original AL algorithm.
%\begin{table}[thb]
	\begin{table}[!t]
	\centering
	
	\begin{tabular}{ccccc}
		\toprule[1pt]
		Algorithm &Soft-DQN &MDQN &AL &clipAL  \\
		\midrule[0.5pt]
		\multirow{2}{*}{Asterix}
		&36.59&12.46&-1.60 &6.90  \\
		&(0.20)&(0.16)&(0.14)&(0.17)\\
		\midrule
		\multirow{2}{*}{Breakout}
		&5.28&63.06&-4.94 &92.34  \\
		&(0.17)&(0.29)&(0.17)&(0.27)\\
		\midrule
		\multirow{2}{*}{Freeway}
		&-3.96&2.03&2.52 &2.10  \\
		&(0.02)&(0.01)&(0.01)&(0.02)\\
		\midrule
		\multirow{2}{*}{Space Invaders}
		&-0.89&20.88&23.02 &27.88  \\
		&(0.11)&(0.22)&(0.16)&(0.18)\\
		\midrule
		\multirow{2}{*}{Seaquest}
		&74.56&136.89&62.75 &100.39  \\
		&(0.25)&(0.41)&(0.64)&(0.36)\\
		\midrule
		\multirow{2}{*}{Pixelcopter}
		&19.52&31.22&39.77 &44.79  \\
		&(0.20)&(0.21)&(0.17)&(0.10)\\
		\midrule
		mean & 21.85 & 44.42 & 20.25  &45.73\\
		\bottomrule[1pt]
	\end{tabular}
	\caption{DQN-normalized score comparison, which represents the percentage (\%) of performance improvement than the DQN baseline. All the results are averaged over 5 seeds, and one standard deviation included in the parenthesis.}
	\label{table:dqn-normalized}
\end{table} 

Naturally, our method can be easily extended to the family of AL-based algorithms, so we also apply our clipping mechanism to two variants of AL-based algorithms, and all the results and analysis are provided in Appendix B.2.2.

\begin{figure}[tb]
	\centering
	\subfigcapskip=0pt
	\subfigure[]{
			\includegraphics[width=0.225\textwidth]{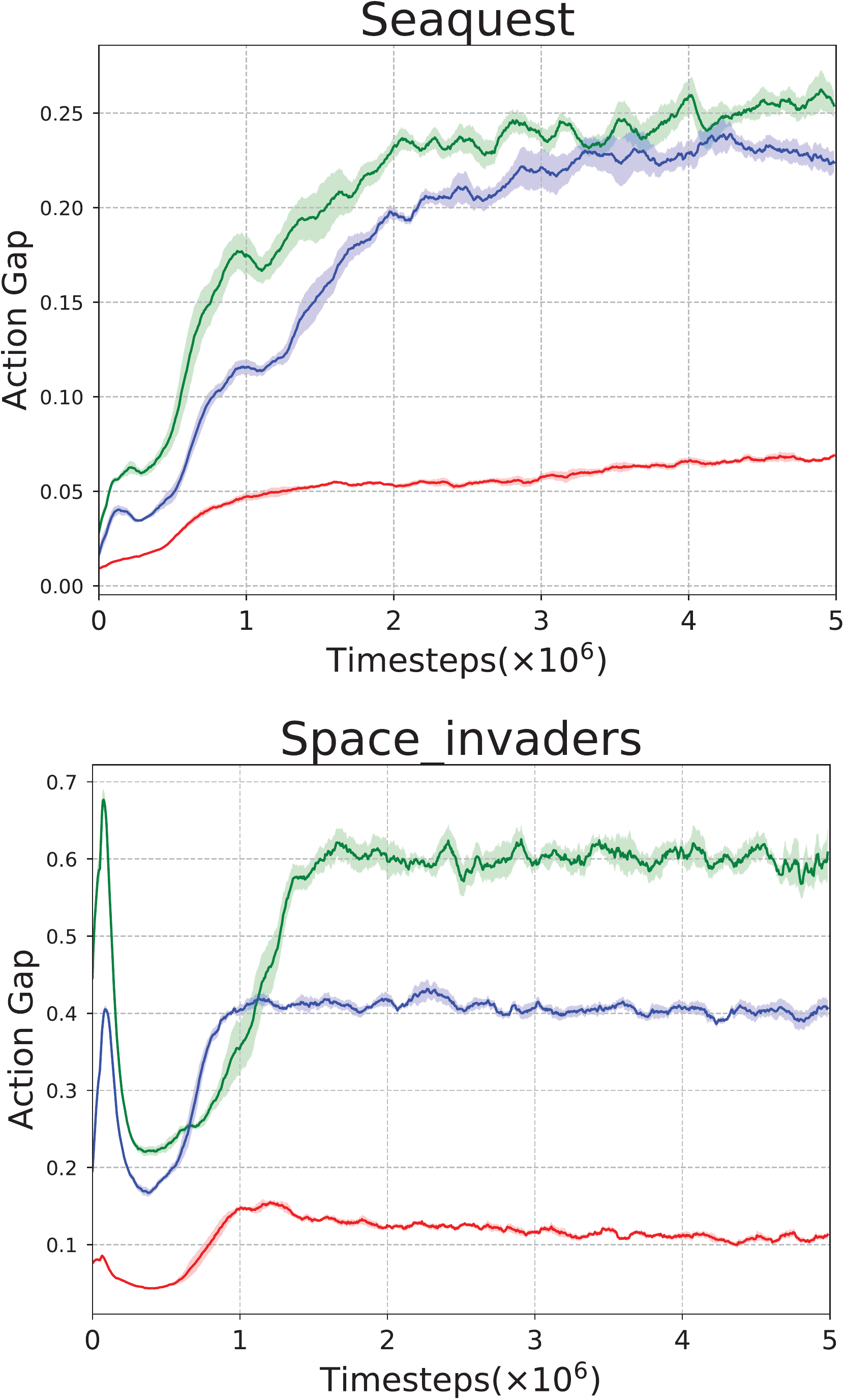}
			\label{fig:action_gap}
	}
	\subfigure[]{
			\includegraphics[width=0.22\textwidth]{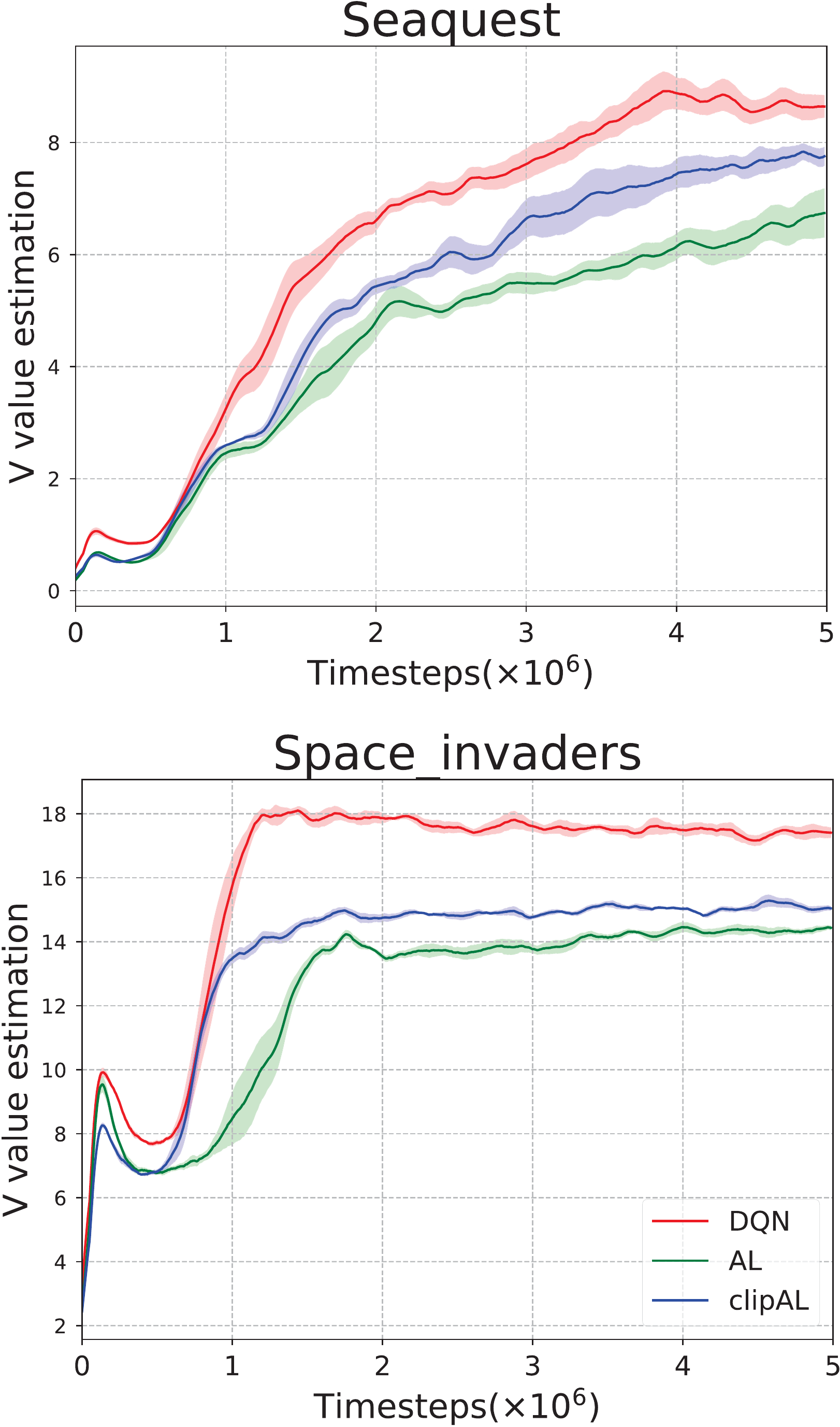}
			\label{fig:mean_v}
	}
	\caption{Comparison about the action gap and the $V$ value estimations on Seaquest and Space-invaders tasks during the training process. Top subfigures represent the results on Seaquest task and the bottom ones are on Space invaders task. \textbf{a)}: action gap estimations; \textbf{b)}: $V$ value estimations.}
	\label{fig:property_analysis} %% label for entire figure
\end{figure}

\subsubsection*{Property Analysis.} As mentioned before, our clipped AL aims to achieve a balance between large action gaps and fast convergence of value function, which is thought of as the main incentive of superior performance. So in this section, we mainly verify whether our clipped AL can achieve this goal.
We estimate the both variables for $\mathcal{T}, \defop{AL}$, and $\defop{clipAL}$ during the training process. Specifically, we denote the $V$ value by $V_\theta(s)=\max_{a}Q_\theta(s,a)$, and the empirical action gap by the difference of estimated values between the best and second best actions: $Q_\theta(s, a^*)-\max_{a\in\mathcal{A}\backslash \left\{a^*\right\}}Q_\theta(s,a)$ with $a^*=\arg\max_{a\in\mathcal{A}}Q_\theta(s,a)$ \cite{vieillard20}.

The results in Figure \ref{fig:property_analysis} include the $V$ value and action gap estimations of DQN, AL, and clipped AL on Seaquest and Space-invaders tasks. Figure \ref{fig:action_gap} shows the estimations of action gaps, in which the action gap of our clipped AL lies between the ones of DQN and AL for the both tasks. These results correspond to our theoretical analysis on the relationship of action gaps in Theorem \ref{them_ag}. And Figure \ref{fig:mean_v} shows the $V$ value estimations of the three algorithms, we can see that the $V$ value of our clipped AL converges faster than AL, in spite of slower than DQN. Combining with the comparisons of both variables, our clipped AL does achieve such a balance between fast convergence and large action gaps, which verifies the feasibility and effectiveness of our motivations.

\subsection{Ablation Study}

According to Theorem \ref{thm_error}, we know that the trade-off between convergence speed and action gap can be achieved by tuning the scaling parameter $\alpha$ and clipping ratio $c$. In this section, we do an ablation study on both critical parameters. We mainly compare the action gap obtained by all the combinations from the candidate set: $\alpha=c=\left\{0.9, 0.8, 0.7\right\}$.

\begin{figure}[tb]
	\centering
	\subfigcapskip=0pt
	\subfigure{
			\includegraphics[width=0.49\textwidth]{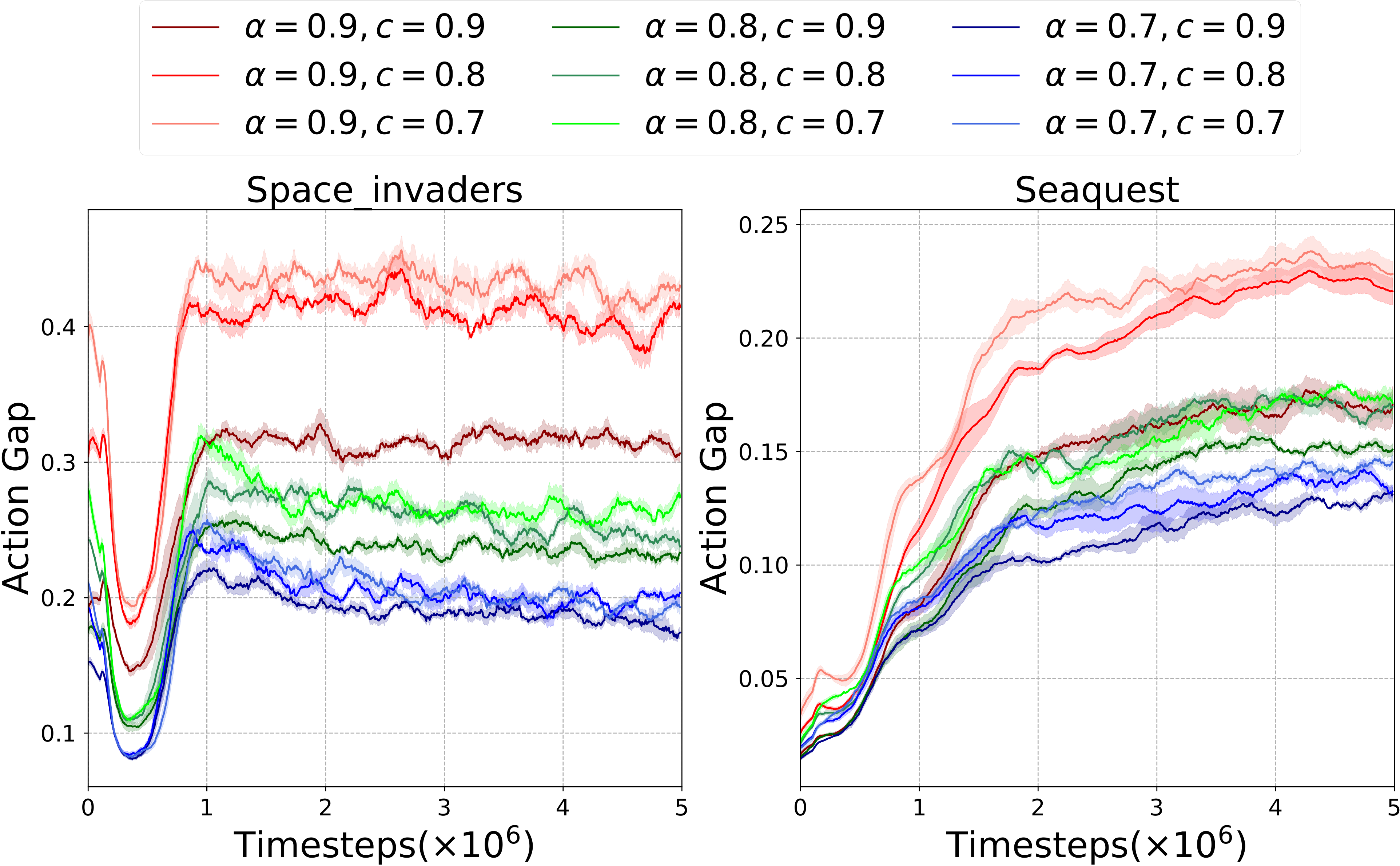}
	}
%	\subfigure[]{
%		\includegraphics[width=0.22\textwidth]{ratio_c/Seaquest-MinAtar-v0_action_gap.pdf}
%		\label{fig:seaquest_ratio_c}
%	}
	\caption{Ablation study on scaling parameter $\alpha$ and clipping ratio $c$ in clipped AL. Comparison about the action gap with different hyperparameter combinations.}
	\label{fig:ablation_study} %% label for entire figure
\end{figure}

Figure \ref{fig:ablation_study} shows the action gap comparison about different parameter combinations on Space invaders and Seaquest tasks. We can see that, when fixing $\alpha$, the action gap will decrease with the increase of clipping ratio $c$; this is intuitive because a larger $c$ means the less probability to add the advantage term, leading to a smaller action gap. While a larger $\alpha$ with a fixed $c$ can lead to more action gaps because $\alpha$ determines the scaling of advantage term, \textit{i.e.,} gap-increasing. In other words, $\alpha$ controls the overall action gaps of all state-action pairs, and our clipping ratio $c$ can further adjust the individual action gap for each state-action pair selectively so as to achieve the finer balance.

\section{Conclusion}

Advantage Learning (AL) is considered to be more robust due to its regularization on the action gap. However, our analysis reveals that AL may cause worse performance loss bound, leading to a slower value convergence if increasing the action gap blindly. In this paper, we propose the clipped AL to adjust the advantage term adaptively so as to increase the action gap more reasonably. This simple modification can obtain better performance with faster convergence while maintaining a proper action gap to keep its robustness and be extended to the family of gap-increasing operators easily. The theoretical and empirical results also confirm the rationality and effectiveness of our proposed methods.

An interesting future study is to design an adaptive clipping ratio $c$ for the training process. Because the clipping mechanism may be more necessary for the robust gap-increasing at the early training stage. While when the induced optimal actions align with the true optimal ones at the late training stage, increasing the action gap for all state-action pairs is more important.

\section*{Acknowledgments}
This work is partially supported by National Science Foundation of China (61732006,61976115), and National Key R\&D program of China (2021ZD0113203). We would also like to thank the anonymous reviewers, for offering thoughtful comments and helpful advice on earlier versions of this work. We also thank Yuhui Wang and Qingyuan Wu for their constructive discussion in the early stage.

\bibliographystyle{aaai22}
\bibliography{aaai22}



\section*{Supplementary material}

# A Theorem Proof

## A.1 Proof of Theorem 1

*Proof.* First, we derive the difference between Bellman operator w.r.t the optimal policy $\pi^*$, *i.e.*, $\mathcal{T}_{\pi^*}$, and the AL operator $\mathcal{T}_{\mathrm{AL}}$ with the same state value function:

$$\begin{aligned}
&\mathcal{T}_{\pi^*} V_k(s) - \mathcal{T}_{\mathrm{AL}} V_k(s) \\
&= r(s, \pi^*(s)) + \gamma \mathbb{E}_{s' \sim P(s'|s, \pi^*(s))} [V_k(s')] - \max_a [\mathcal{T} Q_k(s, a) - \alpha(V_k(s) - Q_k(s, a))] \\
&\leq r(s, \pi^*(s)) + \gamma \mathbb{E}_{s' \sim P(s'|s, \pi^*(s))} [V_k(s')] - \mathcal{T} Q_k(s, \pi^*(s)) + \alpha(V_k(s) - Q_k(s, \pi^*(s))) \\
&= r(s, \pi^*(s)) + \gamma \mathbb{E}_{s' \sim P(s'|s, \pi^*(s))} [V_k(s')] - [r(s, \pi^*(s)) + \gamma \mathbb{E}_{s' \sim P(s'|s, \pi^*(s))} [V_k(s')]] \\
&\quad + \alpha(V_k(s) - Q_k(s, \pi^*(s))) \\
&= \alpha(V_k(s) - Q_k(s, \pi^*(s)))
\end{aligned} \tag{1}$$

We define $\Delta_k^{\pi^*}(s) = V_k(s) - Q_k(s, \pi^*(s))$, and let $\alpha \Delta_k^{\pi^*} = \mathcal{T}_{\pi^*} V_k - \mathcal{T}_{\mathrm{AL}} V_k$ represent the corresponding vector of length $|\mathcal{S}|$ where each entry is $\alpha \Delta_k^{\pi^*}(s)$.

And then we can derive the error bound for $V^* - V_{k+1}$ which denotes the distance of $(k+1)$-th iteration in $\{V_k\}_{k=0}^K$ to the optimal state-value function $V^*$:

$$\begin{aligned}
&V^* - V_{k+1} \\
&= \mathcal{T}_{\pi^*} V^* - \mathcal{T}_{\pi^*} V_k + \mathcal{T}_{\pi^*} V_k - \mathcal{T}_{\mathrm{AL}} V_k \\
&\leq \gamma P_{\pi^*} (V^* - V_k) + \alpha \Delta_k^{\pi^*}
\end{aligned} \tag{2}$$

Where $P_{\pi^*}$ is the probability transition matrix of size $|\mathcal{S}| \times |\mathcal{S}|$, *i.e.*, $P_{\pi^*}(s, s') = P(s'|s, \pi^*(s))$. By recursively applying the Eq.(2) for $K$ times, we have:

$$\begin{aligned}
&V^* - V_K \\
&\leq \gamma P_{\pi^*} (V^* - V_{K-1}) + \alpha \Delta_{K-1}^{\pi^*} \\
&\leq (\gamma P_{\pi^*})^2 (V^* - V_{K-2}) + \gamma P_{\pi^*} \left( \alpha \Delta_{K-2}^{\pi^*} \right) + \alpha \Delta_{K-1}^{\pi^*} \\
&\cdots \\
&\leq (\gamma P_{\pi^*})^K (V^* - V_0) + \sum_{k=0}^{k=K-1} (\gamma P_{\pi^*})^{K-k-1} \left( \alpha \Delta_k^{\pi^*} \right)
\end{aligned} \tag{3}$$

Let us denote the greedy policy with respect to $V_K$, *i.e.*, $\pi_{K+1}$ by $\pi_{K+1}(s) \triangleq \max_a [r(s, a) + \gamma \mathbb{E}_{s' \sim P(s'|s, a)} [V_K(s')]]$, and $V^{\pi_{K+1}}$ denote the fixed point of $\mathcal{T}_{\pi_{K+1}}$, *i.e.*, $V^{\pi_{K+1}} = \mathcal{T}_{\pi_{K+1}} V^{\pi_{K+1}}$. Then we derive the $V^* - V^{\pi_{K+1}}$ by following the similar process provided in ([3]):

$$\begin{aligned}
&V^* - V^{\pi_{K+1}} \\
&= \mathcal{T}_{\pi^*} V^* - \mathcal{T}_{\pi^*} V_K + \mathcal{T}_{\pi^*} V_K - \mathcal{T} V_K + \mathcal{T} V_K - \mathcal{T}_{\pi_{K+1}} V^{\pi_{K+1}} \\
&\leq \gamma P_{\pi^*} (V^* - V_K) + \mathcal{T}_{\pi_{K+1}} V_K - \mathcal{T}_{\pi_{K+1}} V^{\pi_{K+1}} \\
&= \gamma P_{\pi^*} (V^* - V_K) + \gamma P_{\pi_{K+1}} (V_K - V^{\pi_{K+1}})
\end{aligned} \tag{4}$$

The inequality in Eq.(4) is due to the fact that $\mathcal{T} V_K \geq \mathcal{T}_{\pi^*} V_K$, and utilizes the definition of induced greedy policy:

$$\begin{aligned}
\mathcal{T} V_K(s) &= \max_{a \in \mathcal{A}} [r(s, a) + \gamma \mathbb{E}_{s' \sim P(s'|s, a)} [V_K(s')]] \\
&= r(s, \pi_{K+1}(s)) + \gamma \mathbb{E}_{s' \sim P(s'|s, \pi_{K+1}(s))} [V_K(s')] \\
&= \mathcal{T}_{\pi_{K+1}} V_K(s)
\end{aligned} \tag{5}$$

Further, we can rewrite Eq.(4) as:

$$V^* - V^{\pi_{K+1}} \leq \gamma \left( \mathbf{I} - \gamma P_{\pi_{K+1}} \right)^{-1} \left( P_{\pi^*} - P_{\pi_{K+1}} \right) (V^* - V_K) \tag{6}$$

And then plugging Eq.(3) into Eq.(6) and taking $\|\cdot\|_\infty$ on the both sides:

$$\begin{aligned}
&\|V^* - V^{\pi_{K+1}}\|_\infty \\
&\leq \gamma \|(\mathbf{I} - \gamma P_{\pi_{K+1}})^{-1}\|_\infty \cdot \|P_{\pi^*} - P_{\pi_{K+1}}\|_\infty \cdot \|V^* - V_K\|_\infty \\
&\leq \frac{2\gamma}{1-\gamma} \left[ \|(\gamma P_{\pi^*})^K\|_\infty \cdot \|V^* - V_0\|_\infty + \alpha \sum_{k=0}^{k=K-1} \|(\gamma P_{\pi^*})^{K-k-1}\|_\infty \cdot \|\Delta_k^{\pi^*}\|_\infty \right] \\
&= \frac{2\gamma}{1-\gamma} \left[ 2\gamma^K V_{\max} + \alpha \sum_{k=0}^{k=K-1} \gamma^{K-k-1} \|\Delta_k^{\pi^*}\|_\infty \right]
\end{aligned} \quad (7)$$

$\square$

## A.2 Proof of Theorem 2

**Lemma 1.** $\forall Q_1, Q_2$ that satisfy $V^*(s) = \max_a Q_1(s,a) = \max_a Q_2(s,a)$ for all state $s$, the we have the following:

1. if $Q_1 \geq Q_2$, then $\mathcal{T}_{\mathrm{AL}} Q_1 \geq \mathcal{T}_{\mathrm{AL}} Q_2$;

2. if $V^*(s) = Q_1(s, \pi^*(s))$, then $\forall n \in \mathbb{N}^+, V^*(s) = \max_a (\mathcal{T}_{\mathrm{AL}})^n Q_1(s,a)$, and $\mathcal{T}(\mathcal{T}_{\mathrm{AL}})^n Q_1 = Q^*$.

*Proof.* According to the definition of $\mathcal{T}_{\mathrm{AL}}$, we have:

$$\begin{aligned}
&\mathcal{T}_{\mathrm{AL}} Q_1(s,a) - \mathcal{T}_{\mathrm{AL}} Q_2(s,a) \\
&= \mathcal{T} Q_1(s,a) + \alpha(Q_1(s,a) - V^*(s)) - \mathcal{T} Q_2(s,a) - \alpha(Q_2(s,a) - V^*(s)) \\
&= \mathbb{E}_{s'} \left[ \max_{a'} Q_1(s',a') - \max_{a'} Q_2(s',a') \right] + \alpha(Q_1(s,a) - Q_2(s,a)) \\
&= \alpha(Q_1(s,a) - Q_2(s,a))
\end{aligned}$$

So if $Q_1 \geq Q_2$, we have $\mathcal{T}_{\mathrm{AL}} Q_{k_1} \geq \mathcal{T}_{\mathrm{AL}} Q_{k_2}$, the first property is verified.

And we also have that:

$$\begin{aligned}
\max_a \mathcal{T}_{\mathrm{AL}} Q_1(s,a) &= \max_a \left[ r(s,a) + \mathbb{E}[V^*(s')] + \alpha(Q_1(s,a) - V^*(s)) \right] \\
&= \max_a \left[ Q^*(s,a) + \alpha(Q_1(s,a) - V^*(s)) \right]
\end{aligned} \quad (8)$$

Due to the conditions:

$$\max_a Q_1(s,a) = Q_1(s, \pi^*(s)) = V^*(s) \quad (9)$$

We know both $\max_a Q^*(s,a)$ and $\max_a Q_1(s,a)$ have the same maximizer $\pi^*(s)$ and maximum $V^*(s)$, so Eq.(8) satisfies:

$$\max_a \mathcal{T}_{\mathrm{AL}} Q_1(s,a) = \mathcal{T}_{\mathrm{AL}} Q_1(s, \pi^*(s)) = V^*(s)$$

So $\mathcal{T}_{\mathrm{AL}} Q_1(s,a)$ also satisfies the condition in Eq.(9). And by applying Eq.(8) and Eq.(9) repeatedly, we can obtain the final result:

$$V^*(s) = \max_a (\mathcal{T}_{\mathrm{AL}})^n Q_1(s,a) \quad (10)$$

Benefiting from the above result, we can further obtain:

$$\begin{aligned}
\mathcal{T}(\mathcal{T}_{\mathrm{AL}})^n Q_1(s,a) &= r(s,a) + \gamma \mathbb{E}_{s'} \left[ \max_{a'} (\mathcal{T}_{\mathrm{AL}})^n Q_1(s',a') \right] \\
&= r(s,a) + \gamma \mathbb{E}_{s'} [V^*(s)] \\
&= Q^*(s,a)
\end{aligned} \quad (11)$$

$\square$

Define $\{\tilde{Q}_k\}_{k=0}^{\infty}, \{\hat{Q}_k\}_{k=0}^{\infty}$ as the $Q$ value function sequences obtained by $\mathcal{T}_{\text{clipAL}}$ and $\mathcal{T}_{\text{AL}}$ respectively. And the corresponding $V$ value function sequences are defined as $\{\tilde{V}_k\}_{k=0}^{\infty}, \{\hat{V}_k\}_{k=0}^{\infty}$ by following $V(s) = \max_a Q(s, a)$. According to the Theorem 1 and Definition 1 in [2], we know that $\mathcal{T}_{\text{clipAL}}$ and $\mathcal{T}_{\text{AL}}$ are optimality-preserving, and thus both the $V$ value function sequences will converge to the optimal value function, *i.e.* $\lim_{k \to \infty} \tilde{V}_k(s) = \lim_{k \to \infty} \hat{V}_k(s) = V^*(s)$. Now we proof the Theorem 2:

*Proof.* According to the definition of optimality-preserving, we know that there is at least one optimal action remains optimal, and all suboptimal actions remain suboptimal, so the induced greedy policy w.r.t $\tilde{Q}_\infty(s, a)$ must belong to the set of optimal policies $\Pi^*$, and we denote it by $\pi^*$. And thus $\pi^*$ satisfies:

$$V^*(s) = \max_a \tilde{Q}_\infty(s, a) = \tilde{Q}_\infty(s, \pi^*(s)) \tag{12}$$

Firstly, according to the definitions of $\mathcal{T}_{\text{clipAL}}$ and $\mathcal{T}_{\text{AL}}$, we have:

$$\mathcal{T}_{\text{clipAL}} \tilde{Q}_\infty(s, a) \geq \mathcal{T}_{\text{AL}} \tilde{Q}_\infty(s, a) \tag{13}$$

Secondly, benefiting from the second property in Lemma 1, and $\tilde{V}$ has been convergent, *i.e.*, $\max_a (\mathcal{T}_{\text{clipAL}})^n \tilde{Q}_\infty(s, a) = \max_a \tilde{Q}_{\infty+n}(s, a) = V^*(s)$:

$$V^*(s) = \max_a \mathcal{T}_{\text{clipAL}} \tilde{Q}_\infty(s, a) = \max_a \mathcal{T}_{\text{AL}} \tilde{Q}_\infty(s, a) \tag{14}$$

And we see that Eq.(13) and Eq.(14) match the conditions of the first property in Lemma 1, so we have:

$$\mathcal{T}_{\text{AL}} \mathcal{T}_{\text{clipAL}} \tilde{Q}_\infty(s, a) \geq (\mathcal{T}_{\text{AL}})^2 \tilde{Q}_\infty(s, a) \tag{15}$$

Based on the definitions again, there exists $(\mathcal{T}_{\text{clipAL}})^2 \tilde{Q}_\infty(s, a) \geq \mathcal{T}_{\text{AL}} \mathcal{T}_{\text{clipAL}} \tilde{Q}_\infty(s, a)$, and combining with Eq.(15), we have:

$$(\mathcal{T}_{\text{clipAL}})^2 \tilde{Q}_\infty(s, a) \geq (\mathcal{T}_{\text{AL}})^2 \tilde{Q}_\infty(s, a) \tag{16}$$

Finally, repeating the process above for $n$ times, we can get $(\mathcal{T}_{\text{clipAL}})^n \tilde{Q}_\infty(s, a) \geq (\mathcal{T}_{\text{AL}})^n \tilde{Q}_\infty(s, a)$. And we take sup and $n \to \infty$ on both sides and obtain:

$$\limsup_{n \to \infty} (\mathcal{T}_{\text{clipAL}})^n \tilde{Q}_\infty(s, a) \geq \limsup_{n \to \infty} (\mathcal{T}_{\text{AL}})^n \tilde{Q}_\infty(s, a)$$
$$\Rightarrow \limsup_{n \to \infty} \tilde{Q}_n(s, a) \geq \limsup_{n \to \infty} \hat{Q}_n(s, a) \tag{17}$$

According to the definitions of action gap, $G(s, a) \triangleq V^*(s) - Q^*(s, a), G_{\text{AL}}(s, a) \triangleq \liminf_{k \to \infty}[\hat{V}_k(s) - \hat{Q}_k(s, a)], G_{\text{clipAL}}(s, a) \triangleq \liminf_{k \to \infty}[\tilde{V}_k(s) - \tilde{Q}_k(s, a)]$ and Eq.(17), we can obtain:

$$-\liminf_{n \to \infty} \tilde{Q}_n(s, a) \leq -\liminf_{n \to \infty} \hat{Q}_n(s, a)$$
$$\Rightarrow V^*(s) - \liminf_{n \to \infty} \tilde{Q}_n(s, a) \leq V^*(s) - \liminf_{n \to \infty} \hat{Q}_n(s, a)$$
$$\Rightarrow \liminf_{n \to \infty} \left[\tilde{V}_n(s) - \tilde{Q}_n(s, a)\right] \leq \liminf_{n \to \infty} \left[\hat{V}_n(s) - \hat{Q}_n(s, a)\right]$$
$$\Rightarrow G_{\text{clipAL}}(s, a) \leq G_{\text{AL}}(s, a) \tag{18}$$

Because $\mathcal{T}_{\text{clipAL}}$ is also gap-increasing as Theorem 1 in [2] defines, then $G(s, a) \leq G_{\text{clipAL}}(s, a)$.

If we want $G_{\text{clipAL}}(s, a) = G_{\text{AL}}(s, a)$, we need $(\mathcal{T}_{\text{clipAL}})^n \tilde{Q}_\infty(s, a) = (\mathcal{T}_{\text{AL}})^n \tilde{Q}_\infty(s, a), \forall n = 1, 2, \cdots$, *i.e.*,:

$$r_n(s, a) \triangleq \frac{(\mathcal{T}_{\text{clipAL}})^n \tilde{Q}_\infty(s, a) - Q_l}{(\mathcal{T}_{\text{AL}})^n \tilde{Q}_\infty(s, a) - Q_l} = \frac{(\mathcal{T}_{\text{AL}})^n \tilde{Q}_\infty(s, a) - Q_l}{V^*(s) - Q_l} \geq c \quad \forall s \in \mathcal{S}, a \in \mathcal{A}, n \in \mathbb{N}^+ \tag{19}$$

Let us see the sequence $\{(\mathcal{T}_{\text{AL}})^n \tilde{Q}_\infty\}_{n=1}^{\infty}$, when $n = 1$[1]:

$$\mathcal{T}_{\text{AL}} \tilde{Q}_\infty = \mathcal{T} \tilde{Q}_\infty + \alpha(\tilde{Q}_\infty - \max_a \tilde{Q}_\infty) = Q^* + \alpha(\tilde{Q}_\infty - V^*) \tag{20}$$

---
[1] For simplicity, we denote $Q - V$ as a vector in $\mathbb{R}^{|\mathcal{S}||\mathcal{A}| \times 1}$ whose each element is $Q(s, a) - V(s)$

3

and when $n = 2$, combining with 20 and the second property in Lemma 1, we can obtain:

$$(\mathcal{T}_{\text{AL}})^2 \tilde{Q}_\infty = \mathcal{T}(\mathcal{T}_{\text{AL}} \tilde{Q}_\infty) + \alpha(Q^* + \alpha(\tilde{Q}_\infty - V^*) - \max_a \mathcal{T}_{\text{AL}} \tilde{Q}_\infty)$$
$$= Q^* + \alpha(Q^* - V^*) + \alpha^2(\tilde{Q}_\infty - V^*) \tag{21}$$

then $\forall n \in \mathbb{N}^+$, the recursive result can be rewritten:

$$(\mathcal{T}_{\text{AL}})^n \tilde{Q}_\infty = V^* + \sum_{t=0}^{t=n-1} \alpha^t (Q^* - V^*) + \alpha^n (\tilde{Q}_\infty - V^*) \tag{22}$$

Next we compute the difference between $(\mathcal{T}_{\text{AL}})^{n+1} \tilde{Q}_\infty$ and $(\mathcal{T}_{\text{AL}})^n \tilde{Q}_\infty$:

$$(\mathcal{T}_{\text{AL}})^{n+1} \tilde{Q}_\infty - (\mathcal{T}_{\text{AL}})^n \tilde{Q}_\infty$$
$$= \alpha^n (Q^* - V^*) + \alpha^{n+1} (\tilde{Q}_\infty - V^*) - \alpha^n (\tilde{Q}_\infty - V^*)$$
$$= \alpha^n (Q^* - \alpha V^* - (1-\alpha) \tilde{Q}_\infty) \tag{23}$$

Here we utilize the Theorem 1 in [5] that $\lim_{k \to \infty} \hat{Q}_k = \frac{Q^* - \alpha V^*}{1-\alpha}$, then:

$$(\mathcal{T}_{\text{AL}})^{n+1} \tilde{Q}_\infty - (\mathcal{T}_{\text{AL}})^n \tilde{Q}_\infty$$
$$= \alpha^n (Q^* - \alpha V^* - (1-\alpha) \tilde{Q}_\infty)$$
$$= \alpha^n (1-\alpha) (\lim_{k \to \infty} \hat{Q}_k - \tilde{Q}_\infty)$$
$$\leq 0 \tag{24}$$

The final inequality is based Eq 17, so we know that $\{(\mathcal{T}_{\text{AL}})^n \tilde{Q}_\infty\}_{n=1}^\infty$ is a decreasing sequence that converges to $\frac{Q^* - \alpha V^*}{1-\alpha}$. According to the condition in Eq 19, when $Q_l \leq \min_{s,a} \frac{Q^* - V^*}{1-\alpha}$, $r_n(s,a)$ could keep non-negative for all state-action pairs, and in this case, for the goal that $G_{\text{clipAL}}^* = G_{\text{AL}}$, the clipping ratio $c$ should satisfy:

$$c \leq \min_{s,a,n} r_n(s,a) = \min_{s,a,n} \frac{(\mathcal{T}_{\text{AL}})^n \tilde{Q}_\infty(s,a) - Q_l}{V^*(s) - Q_l} = \min_{s,a} \frac{Q^* - \alpha V^* - (1-\alpha) Q_l}{(1-\alpha)(V^* - Q_l)} \tag{25}$$

□

## B  Experiment supplementary

### B.1  Chain-walk Example

In the 11-state chain-walk example, we adopt the tabular update to learn the $Q$ value for all state-action pairs. All the hyperparameters for $\mathcal{T}$, $\mathcal{T}_{\text{AL}}$ and $\mathcal{T}_{\text{clipAL}}$ are chosen as Table 1

Table 1: Hyperparameters used in numerical experiments of 11-states chain-walk

|  | $\gamma$ | $\alpha$ | $c$ |
|---|---|---|---|
| $\mathcal{T}$ | 0.99 | / | / |
| $\mathcal{T}_{\text{AL}}$ | 0.99 | 0.99 | / |
| $\mathcal{T}_{\text{clipAL}}$ | 0.99 | 0.99 | 0.9 |

When evaluating *action-gap deviation* $\kappa$, we need to sample from $\mathcal{S}^+$ to estimate its value. In this paper, we use 1000 samples to compute $\kappa$.

Besides the numerical analysis in the main body, we do ablation study on the clipping ratio $c$ in the chain-walk example. We compare the $V$ value convergence , mean action gap, and action gap deviation $\kappa$ for different choices of $c \in (0,1)$, and the results are shown in Figure 1

We mainly evaluate $\mathcal{T}_{\text{clipAL}}$ with the candidate set $c = \{0.6, 0.8, 0.9, 0.95\}$ and compare their results with $\mathcal{T}$ and $\mathcal{T}_{\text{AL}}$. Figure 1(a) shows the $V$ value convergence results and we can see that, with the

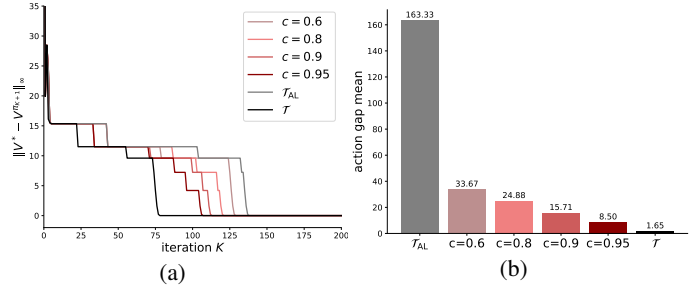

Figure 1: Ablation study on the clipping ratio $c$ for 11-state chain-walk example. **a**): $V$ value convergence comparison; **b**): mean of action gap at iteration 500.

increase of the clipping ratio $c$, $\mathcal{T}_{\text{clipAL}}$ can obtain a faster convergence to the optimal. This result is intuitive because $\mathcal{T}_{\text{clipAL}}$ with a larger $c$ will be closer to $\mathcal{T}$, and otherwise be closer to $\mathcal{T}_{\text{AL}}$. The averaged action gaps after the convergence of $V$ value are shown in Figure 1(b), all the choices of $c$ for $\mathcal{T}_{\text{clipAL}}$ will obtain a middle action gap between the ones by $\mathcal{T}_{\text{AL}}$ and $\mathcal{T}$, and smaller $c$ could lead to a larger gap increasing.

## B.2 MinAtar and PLE

### B.2.1 Clipped AL and baselines.

We implement all the MinAtar and PLE tasks based on the *Explorer* platform [6], and adopt its most hyperparameters configures except its learning rate $\eta$. Instead of the $\eta$ candidate set used in Explorer, we select the official recommendation $\eta = 0.00025$ in MinAtar [8]. As for the specific parameters of all baselines, we adopt the choices given in [7], also shown in Table 2. Our clipped AL introduces two extra parameters $Q_l$ and $c$, specific choices are shown in Table 3.

Table 2: Shared specific parameters for MinAtar and PLE

| MDQN specific parameters | |
|---|---|
| $\tau$ (entropy temperature) | 0.03 |
| $\alpha$ (log-policy scaling term) | 0.9 |
| $l_0$ (clipping value) | -1 |
| AL & PAL specific parameters | |
| $\alpha$ (advantage scaling term) | 0.9 |

Table 3: Learning rate and clipping ratio for MinAtar and PLE

| Parameter | MinAtar | | | PLE | | |
|---|---|---|---|---|---|---|
| | clipAL | clipPAL | clipMDQN | clipAL | clipPAL | clipMDQN |
| $\eta$ | 0.00025 | | | 0.00003 | | |
| $c$ | 0.8 | | | 0.7 | | 0.6 |
| $Q_l$ | 0 | | | -1 | | |

### B.2.2 Extension

It's natural that our clipping mechanism can be extended to the family of AL-based methods. In this section, we introduce two strong variants of AL algorithms and present their enhanced version like our clipped AL.

5

**Persistent AL.** [2] argues that it's beneficial to encourage the *persistence* of the greedy policies, *i.e.,* the infrequent switch between actions, and then proposes the *persistent advantage learning* (PAL) operator based on $\mathcal{T}_{\text{AL}}$:

$$\mathcal{T}_{\text{PAL}}Q(s,a) \triangleq \max\left\{\mathcal{T}_{\text{AL}}Q(s,a), r(s,a) + \gamma \mathbb{E}_{s' \sim P(s'|s,a)}\left[Q(s',a)\right]\right\} \quad (26)$$

Similarly, we apply the same clipping mechanism to the PAL operator, and then define the *Clipped Persistent Advantage Learning* (**clipped PAL**) operator as:

$$\mathcal{T}_{\text{clipPAL}}Q(s,a) \triangleq \max\left\{\mathcal{T}_{\text{clipAL}}Q(s,a), r(s,a) + \gamma \mathbb{E}_{s' \sim P(s'|s,a)}\left[Q(s',a)\right]\right\} \quad (27)$$

**Munchausen DQN.** According to [7], the chosen baseline, MDQN, can also be viewed as a *soft advantage learning* operator and its value update can be rewritten as:

$$Q_{k+1} = r + \gamma P \langle \pi_{k+1}, Q_k - \tau \ln \pi_{k+1}\rangle + \alpha\tau \ln \pi_{k+1} \text{ s.t. } \pi_{k+1} = \arg\max_{\pi}\langle \pi, Q_k\rangle + \tau \mathcal{H}(\pi)$$

$$\Rightarrow Q_{k+1} = r + \gamma P(\tau \ln \langle 1, \exp \frac{Q_k}{\tau}\rangle) + \alpha(Q_k - \tau \ln\langle 1, \exp\frac{Q_k}{\tau}\rangle) \quad (28)$$

When taking $\alpha \to 0$, MDQN is the same with AL. So the additional *log-policy term* $\alpha\tau \ln \pi_{k+1}$ can be seen as a *soft advantage learning term* which helps to increase the action-gap smoothly. We also extend the clipping mechanism to MDQN, and define the **clipped MDQN** which replaces the log-policy term with:

$$\tau \ln \pi(a|s) \cdot \mathbb{I}\left[\frac{Q(s,a) - Q_l}{V(s) - Q_l} \geq c\right] \quad (29)$$

**Empirical Results.** To compare the AL-based methods (AL, PAL and MDQN) and their clipped versions, we conduct experiments on four MinAtar and PLE tasks and show their performance curve in Figure 2. All the specific hyperparameters are given in Table 2 and Table 3.

We can see that our clipping mechanism can still achieve certain performance improvements when applied to PAL and MDQN, though the improvements are not as significant as the ones obtained by the clipped AL. Besides, our methods also keep the best results in Breakout, Space-invaders and Pixelcopters tasks and the clipping mechanism can generally improve the sample efficiency of AL-based method. Due the max operator used in PAL and the clipping function on the log-policy term in MDQN, they can also mitigate the blindness to some extent when implementing gap-increasing, which we think of as the reason why clipped PAL and clipped MDQN not achieve such significant improvement as clipped AL does.

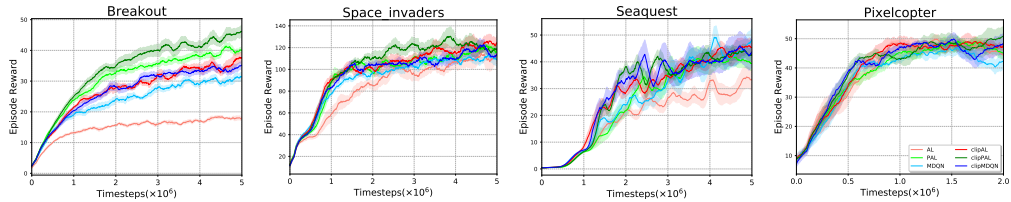

Figure 2: Evaluation episode reward comparison between the AL-based algorithms and their clipped versions on four tasks. All the results are averaged over five random seeds, with the shade area corresponding to one standard error.

### B.3 Atari

We further compare our method on the more complex Atari benchmark [1]. We run our clipped AL and the original AL methods across a subset of Atari tasks (total 12 games). All the implementations are based on the APE-X [4] framework[2], which can accelerate the learning due to a distributed architecture.

---

[2]We use one implementation of APE-X:https://github.com/jingweiz/pytorch-distributed

We mainly compare three algorithms (DQN, AL, clipAL) on these tasks. We select the same scaling coefficient $\alpha = 0.9$ for the both methods. For a convenience, we set the $Q_l = 0$ for all tasks and finetune the clipping ratio from a candidate set $\{0.85, 0.9, 0.95\}$ for each task when implementing our clipped AL. We run $1\mathbf{M}$ training steps for all tasks and conduct a separate evaluation period each $10\mathbf{K}$ training steps. All the evaluation results are averaged over 3 random seeds. A full table of our results is provided in Table 4. And Figure 3 demonstrates the performance curves of parts of tasks.

From the results shown below, We can see that our clipped AL can further improve the performance of the original AL method over most tasks, even though the AL method has already achieved a better performance than DQN. Our method own the best or second best performance over the most chosen tasks. So these results again confirm the effectiveness of our clipping mechanism in the complex environments.

| Env | random | DQN | AL | clipAL |
|---|---|---|---|---|
| Asterix | 210.0 | 5296.67 | 7426.67 | **9997.5** |
| Asteroids | 719.1 | 428.67 | 1281.67 | **1359.17** |
| Beam Rider | 363.9 | 2693.53 | **5706.67** | 4592.83† |
| Bowling | 23.1 | 48.38† | 37.45 | **54.47** |
| Crazy Climber | 10780.5 | 87931.67 | 99556.67 | **115670.0** |
| Gravitar | 173.0 | 440.0 | 398.33 | **661.67** |
| Gopher | 257.6 | 3537.66 | 43241.33 | **45433.67** |
| Kangaroo | 52.0 | 12340.0 | 14090.0 | **14756.67** |
| Ms. Pacman | 307.3 | 3675.0 | **3842.83** | 3764.5† |
| Phoenix | 761.4 | 8229.5 | **8665.5** | 8328.83† |
| Q*bert | 163.0 | 1087.08 | 1439.58 | **2067.08** |
| Seaquest | 68.4 | **2702.0** | 1522.67 | 2018.67† |

Table 4: Performance comparison of a subset of Atari tasks in final evaluation. All the results are averaged over 3 random seeds. The **bold** values represent the best results, and † means the second best results.

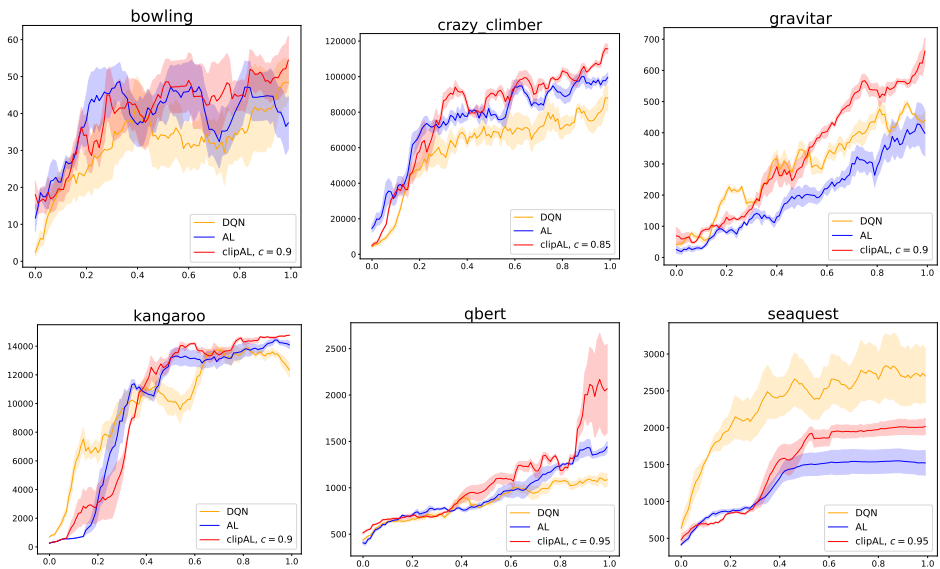

Figure 3: Performance comparison between AL and our clipAL on a subset of Atari tasks. The $x$-axis represents the training steps (total $1\mathbf{M}$) and the $y$-axis indicates the evaluation performance (undiscounted episode reward). All the results are averaged over 3 random seeds, with the shade area corresponding to one standard error.

7



\end{document}